\documentclass[lettersize,journal]{IEEEtran}
\usepackage{amsmath,amsfonts}
\usepackage{algorithmic}
\usepackage{algorithm}
\usepackage{array}
\usepackage[caption=false,font=normalsize,labelfont=sf,textfont=sf]{subfig}
\usepackage{textcomp}
\usepackage{stfloats}
\usepackage{url}
\usepackage{verbatim}
\usepackage{graphicx}
\usepackage{cite}
\usepackage{multirow}
\begin{document}

\title{Rethinking Rotation-Invariant Recognition of Fine-grained Shapes from the Perspective of Contour Points}

\author{
Yanjie Xu,  Handing Xu, Tianmu Wang, Yaguan Li, Yunzhi Chen and Zhenguo Nie*

\thanks{This work was supported by the National Key Research and Development Program of China under Grant 2022YFB4703000.}

\thanks{*Corresponding author: Zhenguo Nie.}

\thanks{Yanjie Xu, Handing Xu, Tianmu Wang, Yunzhi Chen and Zhenguo Nie are with the Department of Mechanical Engineering, Tsinghua University, Beijing, 100084, China, and with the State Key Laboratory of Tribology in Advanced Equipment, Tsinghua University, Beijing, 100084, China (e-mail: xuyanjie22@mails.tsinghua.edu.cn; xhd21@mails.tsinghua.edu.cn; wtm21@mails.tsinghua.edu.cn; yunzhichen@tsinghua.edu.cn; zhenguonie@tsinghua.edu.cn;).}

\thanks{Yaguan Li is with the College of Mechanical Engineering, Taiyuan University of Technology, Taiyuan, Shanxi 030024, China(e-mail: liyaguan0014@link.tyut.edu.cn).}

}

\markboth{Journal of \LaTeX\ Class Files,~Vol.~14, No.~8, August~2021}%
{Shell \MakeLowercase{\textit{et al.}}: A Sample Article Using IEEEtran.cls for IEEE Journals}


\maketitle

\begin{abstract}
Rotation-invariant recognition of shapes is a common challenge in computer vision. Recent approaches have significantly improved the accuracy of rotation-invariant recognition by encoding the rotational invariance of shapes as hand-crafted image features and introducing deep neural networks. However, the methods based on pixels have too much redundant information, and the critical geometric information is prone to early leakage, resulting in weak rotation-invariant recognition of fine-grained shapes. In this paper, we reconsider the shape recognition problem from the perspective of contour points rather than pixels. We propose an anti-noise rotation-invariant convolution module based on contour geometric aware for fine-grained shape recognition. The module divides the shape contour into multiple local geometric regions(LGA), where we implement finer-grained rotation-invariant coding in terms of point topological relations. We provide a deep network composed of five such cascaded modules for classification and retrieval experiments. The results show that our method exhibits excellent performance in rotation-invariant recognition of fine-grained shapes. In addition, we demonstrate that our method is robust to contour noise and the rotation centers. The source code is available at \url{https://github.com/zhenguonie/ANRICN_CGA}

\end{abstract}


\section{Introduction}
\IEEEPARstart{O}{ne} important basis for the human brain to distinguish objects is the perception of geometric shapes\cite{Attneave1954}. Shape recognition is not affected by grayscale, color, and texture, which have widely been used in computer vision. Research on shape recognition has made remarkable progress in areas such as object recognition\cite{Wei2023}, pose estimation\cite{Taeyeop2021, Li2021}, and image retrieval\cite{Florian2021}. In recent years, convolutional neural networks(CNNs) have emerged as a simple yet effective solution for shape recognition\cite{Bansal2020}. However, rotation-invariant recognition of shapes is a pervasive challenge, given that CNN lacks rotation invariance\cite{Mei2023}.

\begin{figure}[!t]
\centering
\includegraphics[width=8.8cm]{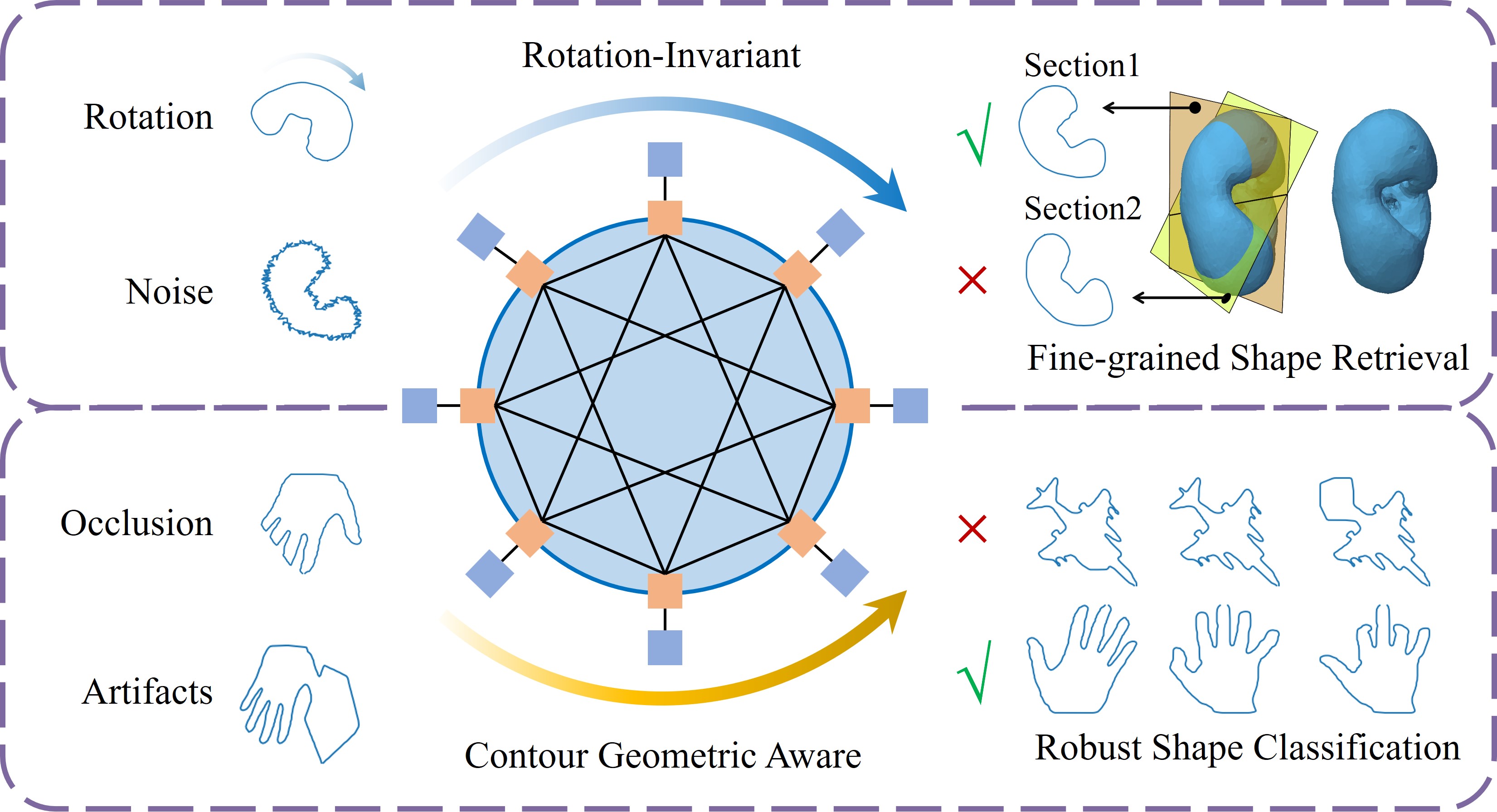}
\caption{The process and tricky scenario of shape recognition}
\label{fig_1}
\end{figure}

A straightforward solution for rotation-invariant recognition is to train CNNs by incorporating rotated shapes, that is data augmentation\cite{Zeiler2014,Bai2020,Bergmann2023}. However, this approach can lead to significant computational costs if we aim to achieve invariant recognition for arbitrary rotation angles. Recently, researchers have developed specific network architectures that make neural networks inherently resistant to rotation\cite{jaderberg2015,Cohen2016,Worrall2017}. However, most of the methods are effective only in the vicinity of a specific rotation angle. Others have been trained by encoding blocks of images into a rotation-invariant format with no loss of shape features\cite{Lucas2024,riccnn2024}. Some recent work has achieved good recognition results without additional data enhancement\cite{riccnn2024}. However, we find that current methods have limitations in the recognition of fine-grained shapes. For example, in some recent work on 2D-3D shape matching\cite{Mohseni2019,Hori2022,Bosma2023,Chang2024,zheng2024}, the small differences in the shape contours of objects greatly increase the difficulty of rotation-invariant recognition. In this fine-grained shape classification and retrieval task, all of our known methods fail to achieve satisfactory results. In addition, boundary noise of shapes is also very tricky. We summarize this tricky scenario as shown in Fig.\ref{fig_1}, where we need a noise-resistant and rotation-invariant recognition method for fine-grained shapes. In computer vision, for the recognized shapes or objects, apart from the texture or light intensity information, their contour shapes are also basic elements for distinguishing them. When we observe a shape, we usually perceive it based on its geometric shape, regardless of how the shape is rotated. For example, we can clearly distinguish a palm from an airplane without seeing the texture. Moreover, our perception of such boundary geometric features is usually sensitive. We can easily capture minor changes on the boundary without affecting our judgment of the whole object. It is not an easy task to enable a deep convolutional neural network to have this ability. In our opinion, both grayscale and binarized images introduce a lot of invalid information in the neural networks except for the contours, which will lose some detail during the convolution and pooling operations. Inspired by work on point cloud recognition\cite{pointnet2017,Chen2022, zhang2022riconv2}, we rethink the shape recognition task in terms of point representations by viewing the shape as a sequence of point sets. We focus only on the geometric features of the contour in the neural networks and achieve feature enhancement through multi-channel convolution to achieve high recognition capability.

In this paper, we propose an anti-noise rotation-invariant convolution module to achieve shape rotation invariance and anti-noise feature encoding. We propose a more stable rotation-invariant feature encoding based on the ordering of contour points, called local orientable axis information(LOAI). The LOAI is a kind of rotation-invariant local reference frame, which is based on contour geometirc aware. We divide the contour into many small segments, each of which is called a local geometric area(LGA). We encode each LGA as a LOAI who stores a portion of the rotation-invariant geometric information of the shape. All the LOAIs are connected and enhanced by the convolutional module, and then fused into a rotation-invariant feature map of the shape. The module can be simply embedded in the deep learning framework. And we provide a neural network consisting of five layers of this module for classification and retrieval. Extensive experimental results in this paper show that the proposed module excellently enhances the accuracy and robustness of neural networks for shape rotation-invariant recognition, especially for fine-grained shapes.

The contributions of this paper include:

1. We propose an anti-noise rotation-invariant convolution module based on contour geometric aware for fine-grained shape retrieval. The module can be simply embedded into the neural network design. And it exhibits excellent performance in rotation-invariant recognition of fine-grained images.

2. We design a local orientable axis(LOA) to encode the rotation-invariant feature of the shape contour. This method augments the capacity of extracting the rotation-invariant features, allowing for a higher recognition rate with fewer augmented samples.

3. Extensive experiments on two classification datasets and four retrieval datasets demonstrate that our method is advanced in fine-grained shape recognition. Additionally, our method is robust to contour noise and the rotation center.

\section{Related Works}

Regarding the rotation-invariant representation for shape recognition, we can classify it into two types from the perspective of data types: methods based on image information and methods based on point information. This section mainly introduces the relevant progress of these two types of methods.

\subsection{Methods based on image information}

In the past few decades, researchers have designed many handcrafted rotation-invariant features(RIFs) based on image information, such as Local Binary Patterns (LBP)\cite{Ojala2002}, Rotation Independent Feature Transform (RIFT)\cite{Lazebnik2005}, Speeded up Robust Features (SURF)\cite{Herbert2008}, etc. With the continuous development of deep learning, handcrafted RIFs are gradually integrated into neural networks. The proposal of spatial transformer networks(STN)\cite{jaderberg2015} is an important work in using learning-based methods to solve the problem of image rotation invariance. Inspired by the SIFT operator\cite{rift2004}, it introduces a learnable spatial transformer module(ST) before the convolution operation. This module improves the network's robustness to the rotation of input images by learning the rotation direction of feature maps. Unfortunately, some recent studies have shown that the learning of intermediate-layer features by the ST module may be inappropriate\cite{Fan2012, Finnveden2020}. Similar to the SIFI of learning image rotation patterns, LBP is a more robust and faster-computing descriptor. RI-LBD defines the local patches of each image as a rotation pattern through a learnable rotation binary pattern encoding\cite{Duan2017}. This encoding method can project rotated images onto the same binary descriptor for clustering. However, this method is highly sensitive to information such as texture noise, and the learning among local image blocks is independent. TBLD preserves the adjacent structure of data through contrastive loss, thereby improving the robustness of rotation-invariant recognition\cite{Miao2021}. The same is true for many methods based on SURF\cite{Hao2024, Maurya2024, Kaur2024}. They all utilize the directional features between rotated samples and reference samples and aim to learn them. The advantage of this approach is that the migration application of the dataset is convenient, but the training cost is high due to the increase in a large number of rotated samples during training. More importantly, it is difficult for this type of method to achieve rotation-invariant recognition at any angle.

Moreover, rotation-invariant feature extraction based on filters is also a commonly used and effective method. Harmonic Networks\cite{hnet2017} apply multiple sets of circular harmonic filters to rotated images and achieve rotation-invariant recognition by learning their maximum filtering responses. This learning mode gets rid of the requirement for input rotated samples and only needs to learn the parameters of the built-in rotation filters to achieve stable recognition. The proposal of Gabor CNN (GCN)\cite{gcn2018} has effectively improved the problem of large-scale rotation recognition. TI-GCN optimizes the structure of Gabor and convolution operations, making the inference speed of deep convolutional neural networks faster\cite{tigcn2020}. However, since the linear combination of directional filters is discrete, there is a range limitation for the recognition of rotation angles. 

In many cases, the rotation angle of the input shapes cannot be predicted in advance. The rotation-invariant features of RIFT for images are achieved by performing recognition on the input image in a rotation-invariant coordinate system. Recently, RIC-CNN introduce this coordinate system into CNN to realize rotation-invariant recognition at any angle without additional data augmentation\cite{riccnn2024}. In addition to the above-mentioned methods, many new ideas have also demonstrated amazing capabilities in the field of shape rotation-invariant recognition recently\cite{e2cnn2021,Mo2024,Li2024}.

\subsection{Methods based on point information}

The methods based on image information usually process the local or overall grayscale information or RGB information of images, and few studies encode the position information of their boundaries. In recent work, Low complex Arrays of Contour Signature (LACS)\cite{Florian2021} proposed a rotation-invariant compact representation by modifying TCD descriptor, which can achieve fast shape indexing in a constant time. Moreover, Complex Representation for Shape Analysis(CRSA)\cite{crsa2024} connects contour points to establish a Topological Feature Map (TFM). This method only considers the topological geometric relationships between the positions of contour pixels, and thus achieves good rotation-invariant recognition accuracy using a RNN. Additionally, some scholars have achieved amazing results in cartoon character recognition by using a transformer encoder to learn from contour sequences\cite{Jia2024}. 

In 3D point cloud analysis, it is common to use position information as the input of the network. For example, Local reference frame(LRF) is an effective way to encode the local geometric features of 3D objects\cite{Kadam2022, Rautek2024}. If considering discrete geometric boundaries as point clouds, we can draw inspiration from the study of rotational invariance of 3D point clouds. In the latest 3D point cloud rotation invariance research\cite{Fei2024}, RIConv++ and PaRIConv\cite{Chen2022, zhang2022riconv2} achieved optimal results in rotation invariant recognition tasks by introducing internal geometric information between local point clouds. However, the geometric topological relationships between 3D local point clouds cannot be applied to shape contours.

Therefore, based on the above analysis, we rethinking the problem of shape rotation-invariant recognition from the perspective of geometric features between contour points, and aim to enable the deep neural network to have a powerful and robust learning representation of shapes.

\begin{figure*}[!t]
\centering
\includegraphics[width=17.2cm]{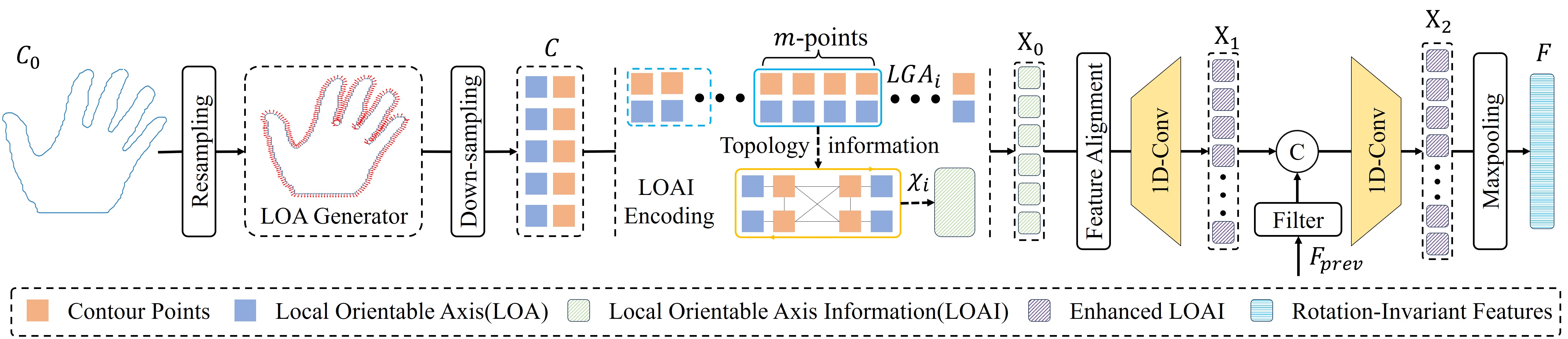}
\caption{The overview of anti-noise rotation-invariant convolution module. }
\label{fig_2}
\end{figure*}

\section{Method}
In this section, we first introduce the overview of the proposed module. Then we demonstrate the encoding of rotation-invariant features based on the local orientable axis information (LOAI) of shape contour points. And we depict the feature enhancement method, which is the key to introduce the proposed module into neural networks. Finally, we provide an effective way to embed the modules into neural networks for shape recognition, which allows for capturing both local and global information of contours.
\subsection{Overview}
In the proposed module, our aim is to encode the input shapes as rotation-invariant features. The Fig.\ref{fig_2} depicts the overall flow of the module. We take as input $C_0$ the contour points of the shape, which can usually be realized by contour extraction or by intercepting the 3D mesh model. We require $C_0$ to be a sequence of points sets along the contour direction. In general, the intervals between adjacent points in $C_0$ are non-uniform. To better capture the geometric information of the shape, we first resample $C_0$ at equal intervals. Then, for each point of the resampled contour points, we calculate its Local Orientable Axis(LOA). The LOA is a vector related to the normal of the contour point, which will be introduced in the subsequent section. The points in $C_0$ with LOA are downsampled to obtain $C$, which is to reduce the computational complexity. Next, we divide $C$ into $k$ local geometric areas(LGAs). Each LGA contains $m$ adjacent points. Consequently, $LGA_i$ encompasses the topology information in the vicinity of the $i$-th LGA. Based on the local reference frame(LRF) and LOA, we encode the topology information within $LGA_i$ as $\chi_i$. And, $k$ LGAs are assembled into $X_0$. We call this process LOAI Encoding. Note that although $X_0$ obtained by the LRF-based encoding method is rotation invariant, it is sensitive to contour noise and the starting point of the sequence $C_0$. Therefore, we add a feature enhancement process which is implemented by two layers of multi-channel 1D convolutional kernels. After the feature enhancement of $X_0$, $X_1$ is obtained. Considering the portability of this module, we allow the enhanced feature $X_1$ to concatenate rotation-invariant features from previous layers, so as to explore deeper feature representations. The concatenated features are dimension-reduced and fused through a multi-channel 1D convolution to obtain $X_2$. Finally, the rotation-invariant feature F of the shape is output through a max-pooling operation. It is worth noting that both the feature enhancement of $X_0$ and the concatenation of $F_{prev}$ require feature alignment before convolution, where $F_{prev}$ is implemented by a more complex filter.

\subsection{LOA Generator}

Local Reference Frame(LRF) is a canonical rotation-invariant coordinate system, but it is noise sensitive. Recently, the LRA proposed in RIConv++\cite{zhang2022riconv2} and the PPF proposed in PaRIConv\cite{Chen2022} show that the normal information can improve the noise-resistance ability. Therefore, we follow the calculation approach of LRA. However, in order to prevent the ambiguity problem of the normal direction caused by covariance calculation, we propose a calculation method named Local Orientable Axis(LOA). This can avoid feature leakage resulting from oscillations in the normal calculation at similar boundaries when encoding fine-grained shapes. In this section, we explain the calculation process. 

\begin{figure}[!h]
\centering
\includegraphics[width=8.0cm]{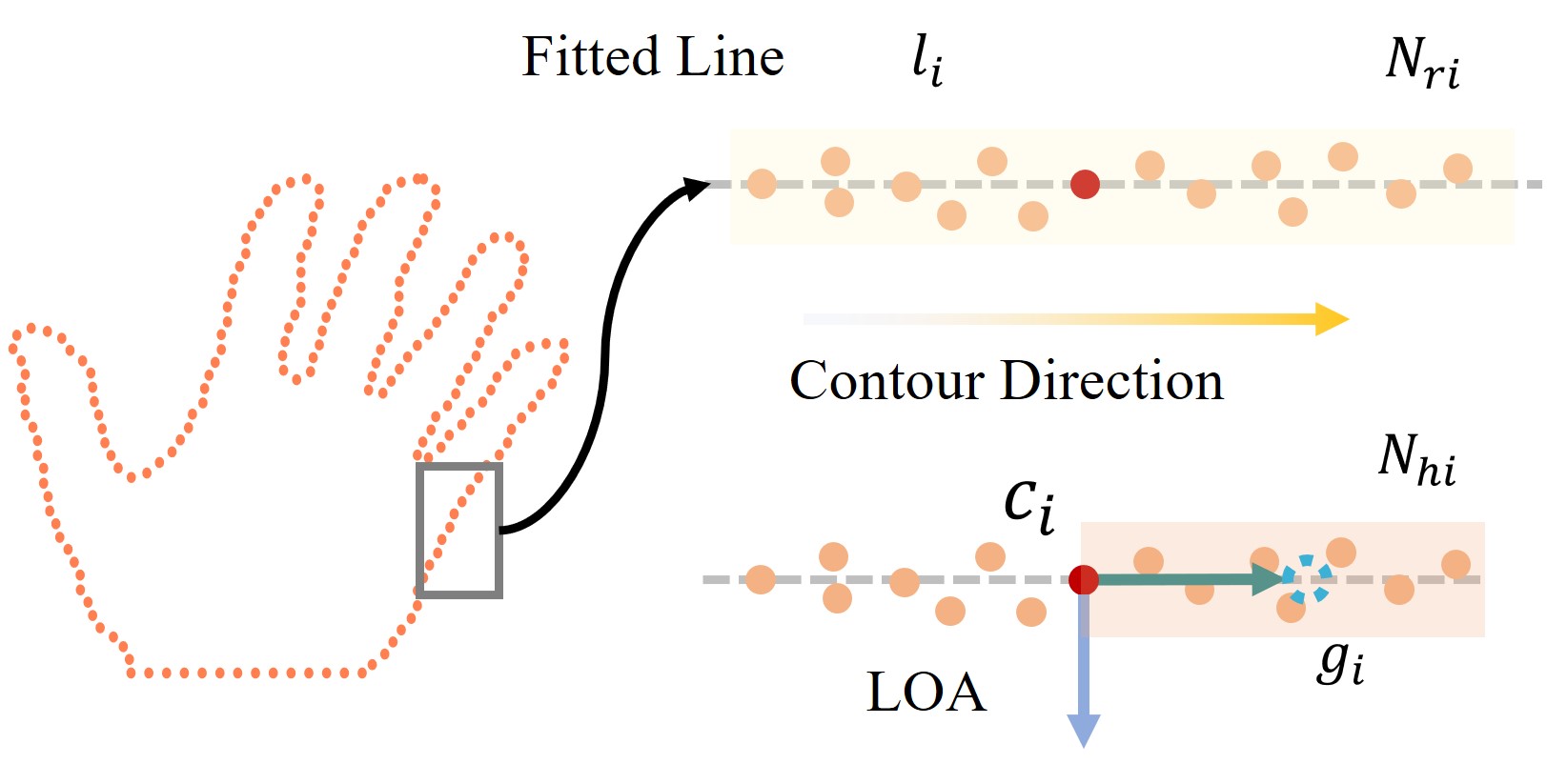}
\caption{The calculation process of the Local Orientable Axis(LOA).}
\label{fig_3}
\end{figure}

Fig.\ref{fig_3} depicts the calculation process of LOA. For the contour point $c_i$, we select $m$ points along its sequence. These points are near $c_i$ and we denote them as $N_{ri}$. Based on the line $l_i$ fitted from $N_ri$, the normal direction of point $c_i$ can be calculated. Simply, we can archieve this through eigenvalue decomposition of the covariance matrix. The normal vector $n_i$ of the point $c_i$ corresponds to the eigenvector associated with the minimum eigenvalue of the covariance matrix derived from the neighborhood point set:

\begin{equation}
M = \sum\limits_{x_i\in N_{ri}}w_i(x_i-c_i)(x_i-c_i)^T
\end{equation}

\noindent where, $N_{ri}$ represent the neighborhood points of $c_i$, and $w_i$ is:

\begin{equation}
w_i = \frac{p-||x_i-c_i||}{\sum_{x_i\in N_s}(p-||x_i-c_i||)}
\end{equation}

\noindent where, $p=max_{x_i\in N_s}(||x_i-P||)$. This weight allows points closer to $c_i$ to contribute more to the normal estimation. For the neighborhood points $N_{ri}$ of the contour sequence, it is easy to obtain the point set $N_{hi}$ located in front of $c_i$. In this paper, We attempt to unify the normal direction by calculating the sign of the angle between the estimated normal vector $n_i$ and the contour direction:

\begin{equation}
LOA_i = sgn(n_i \times (g_i-c_i)))\cdot n_i
\end{equation}

\noindent where $g_i$ is the centorid of $N_{hi}$. And $sgn(\cdot)$ is a sign function:

\begin{equation}
sgn(x)=\left\{
	\begin{aligned}
	-1 \quad x<0\\
	0 \quad x=0\\
	1 \quad x>0\\
	\end{aligned}
	\right
	.
\end{equation}

We name the oriented normal vector as Local Orientable Axis(LOA). Although different contour sequence directions uniformly change the direction of the normal vector, we have experimentally test that this will not affect the recognition accuracy. However, the oriented or unoriented of normal vector has a significant impact on the recognition accuracy, especially for fine-grained shape recognition. We further verify this situation in the discussion section.

\subsection{LOAI Encoding}

\begin{figure}[!h]
\centering
\includegraphics[width=8.8cm]{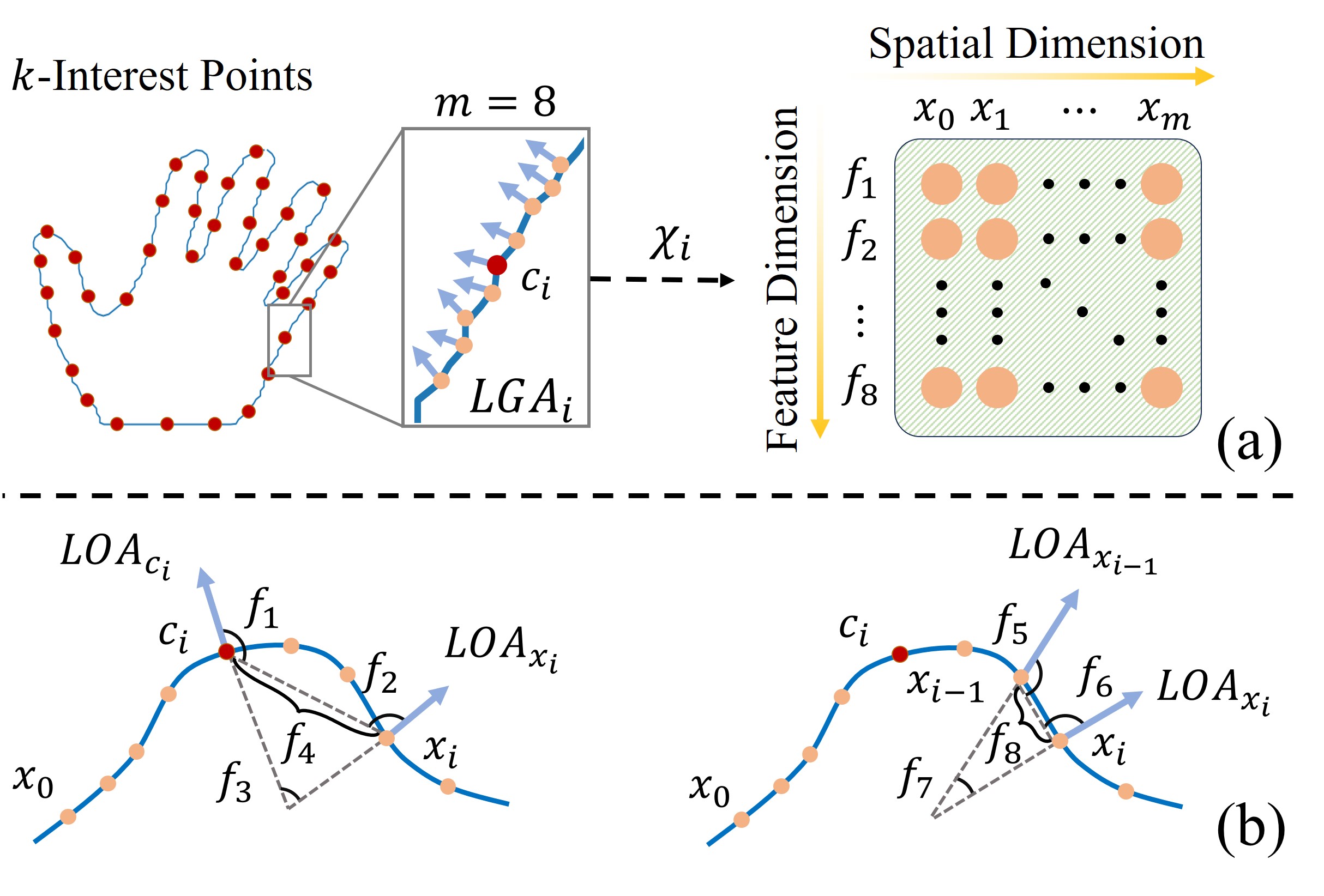}
\caption{The encoding process of Local Orientable Axis Information(LOAI). (a) Local geometric areas are encoded into LOAI matrices. (b) Features of local geometric areas.}
\label{fig_4}
\end{figure}

In order to effectively construct the rotation-invariant features of contour shapes, methods such as FPFH\cite{fpfh2009rusu}, SHOT\cite{shot2017samuele}, and RoPS\cite{RoPS2013Guo} provide effective support. For contour points, the topological connection information between points contains more rotation-invariant features\cite{Lucas2024}. Inspired by the above works, we select $k$ interest points at equal distances on the contour, and focus on the topology information of $m$ neighboring points centered on each interest point. We name this neighboring region Local Geometric Area(LGA), and there are $k$ LGAs in total. Fig.\ref{fig_4}(a) illustrates the encoding process of $LGA_i$ at a certain interest point $c_i$. For two-dimensional point set sequences, angles and distances are the easiest elements to calculate. The angle information is not affected by shape scaling, but the distance information is highly dependent on the shape ratio. Considering more complex indexing tasks and fine-grained recognition problems, in this paper, we mix angle and distance encoding. As shown in Fig.\ref{fig_4}(b), $f_1, f_2, f_3, f_4$ are commonly used LRF feature operators. A topological triangle can be formed between each neighbor point $x_i$ in the LGA and the interest point $c_i$,  and we encode it as follows:

\begin{equation}
\begin{aligned}
&f_1 = cos(\angle(LOA_{c_i}, x_i-c_i)) \\
&f_2 = cos(\angle(LOA_{x_i}, c_i-x_i)) \\ 
&f_3 = cos(\angle(LOA_{c_i},LOA_{x_i})) \\
&f_4 = ||x_i-c_i|| \\
\end{aligned}
\end{equation}

In recent work\cite{zhang2022riconv2,Chen2022}, research has shown that introducing geometric information between adjacent points other than interest points can improve the recognition ability of neural networks. Therefore, we encode the topological information of the triangle formed between each nerghbor point $x_i$ and its previous point $x_{i-1}$ as the features $f_5,f_6,f_7,f_8$. The calculation of these 4 feature operators is shown in the following formula:

\begin{equation}
\begin{aligned}
&f_5 = cos(\angle(LOA_{x_{i-1}},x_i-x_{i-1})) \\
&f_6 = cos(\angle(LOA_{x_{i}},x_{i-1}-x_i)) \\
&f_7 = cos(\angle(LOA_{x_i},LOA_{x_{i-1}})) \\
&f_8 = ||x_i-x_{i-1}||
\end{aligned}
\end{equation}

\noindent where the previous point of $x_0$ is regarded as $x_7$. Therefore, a point $x_i$ can be encoded into 8 features. As shown in Fig.\ref{fig_4}(a), we combine the features of $m$ points into a matrix $\chi_i$, which is considered as an Local Orientable Axis Information(LOAI):

\begin{equation}
\label{eq_feature}
\chi_i = (F_{x_0},F_{x_1},...,F_{x_m})
\end{equation}

\noindent where $F_{xi} = (f_1,f_2,...,f_8)$. We encode the $k$ LGAs respectively, and these are combined into a feature representation $X_0$ of this shape.

\subsection{Feature enhancement and fusion}

\begin{figure}[h]
\centering
\includegraphics[width=8.8cm]{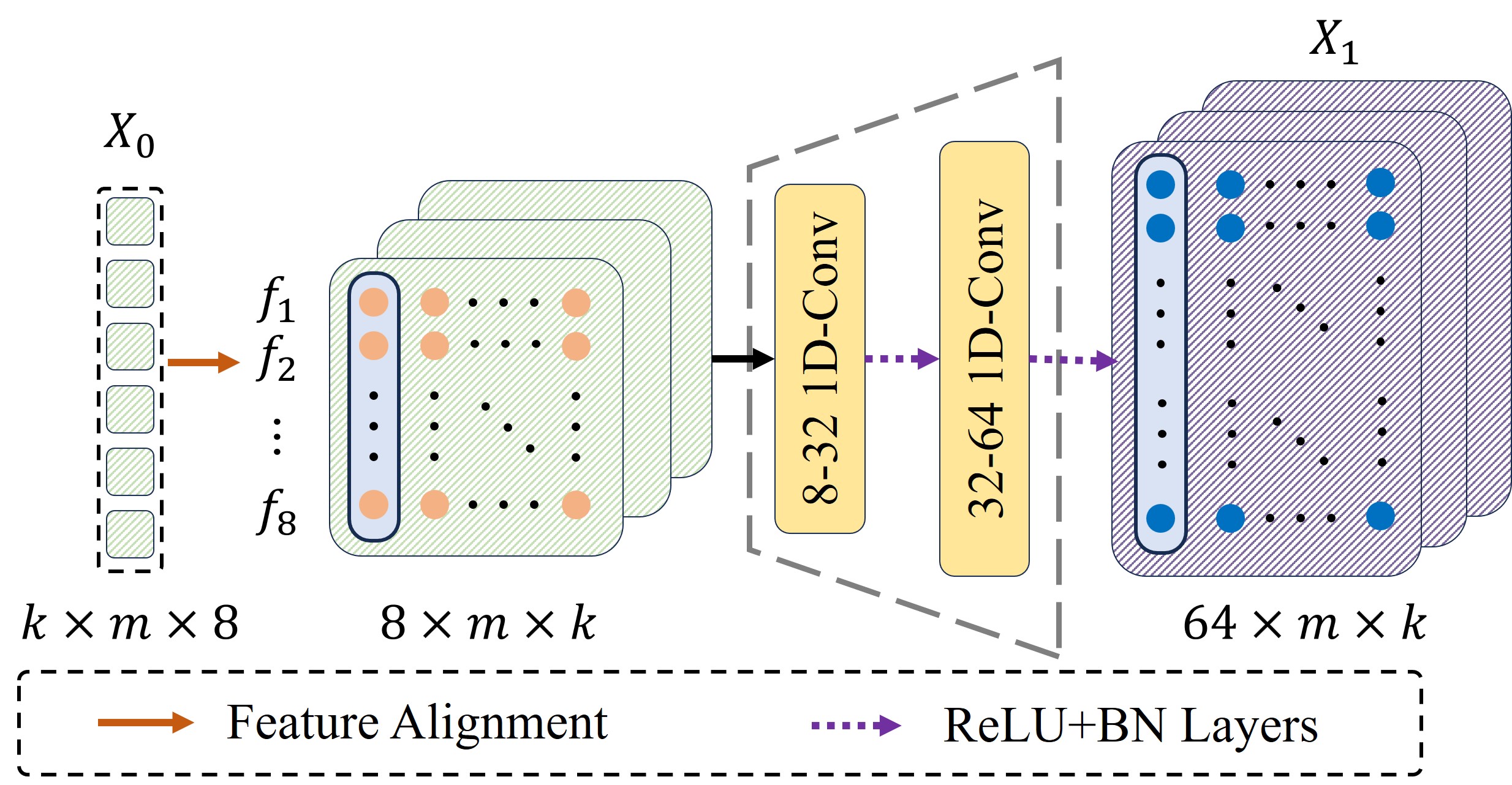}
\caption{Feature Enhancement Module. After the encoded LOAI matrices are combined and aligned, they pass through a two-layer multi-channel 1D convolutional kernel to achieve feature enhancement.}
\label{fig_5}
\end{figure}

In Fig.\ref{fig_2}, we refer to the process from $X_0$ to $X_1$ as feature enhancement. The feature enhancement process has two main objectives: one is to improve the ability to abstract latent features, and the other is to assign weights for training. In fact, the encoded $X_0$ already contains sufficient geometric information of the shape , and these encodings are rotation-invariant. However, the information in $X_0$ is overfitted. When facing shape recognition with different levels of fine-granularity, we hope it can adaptively encode to a degree where clear clustering is achievable. Fortunately, the Eq.\ref{eq_feature} is fully differentiable, which means we can achieve the goal of learning through backpropagation by assigning weights. Therefore, as shown in Fig.\ref{fig_5}, we use two-layer multi-channels 1D-convolutional-kernels to assign weights and mine features for $X_0$. To ensure that the feature mining by the convolution operation is accurate, we need to align the data of $X_0$ so that the convolution is performed along the feature dimension.

In addition, the process from $X_1$ to $F$ can be viewed as feature fusion. Because the module can be embedded in a deep network, concatenation can introduce the information from other layers to further mine potential features. The enhanced feature $X_1$ is combined with the feature $F_{prev}$ from the previous layer. Then the fused features is compressed by a one-layer multi-channel 1D convolution operation to obtain $X_2$. Finally, we select the maximum value on each feature dimension by a max-pooling operation to obtain the rotation-invariant feature $F$.

\subsection{Rotation-invariant recognition network}

\begin{figure*}[!t]
\centering
\includegraphics[width=17.2cm]{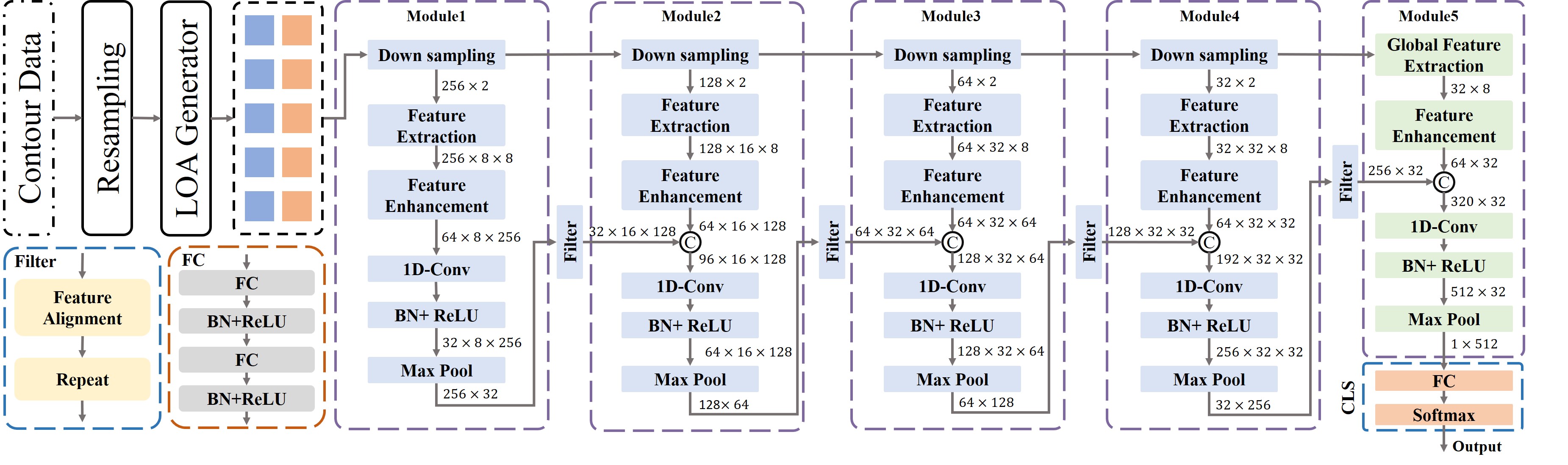}
\caption{The framework of a rotation-invariant recognition network. This framework is composed of five proposed modules connected in series. The contour data is formatted to the input size by the Resampling. The LOA Generator provides the LOA for each contour point. After combining the coordinates of contour points with the LOA, the data undergoes encoding through the five proposed modules, and finally, the output is generated via a Classification Head.}
\label{fig_6}
\end{figure*}

In the proposed module, the calculation process from $C$ to $F$ is fully differentiable, so it can be embedded into a deep neural network to achieve feature learning. In this section, we present a shape rotation-invariant recognition network framework for subsequent experiments. This framework consists of five cascaded encoding modules and ralizes shape recognition through a classification head. As shown in Figure 6, the contour point sequence is fed into the first encoding module after passing through the LOA Generator. 

\begin{table}[h]
\caption{Hyperparametric design for Modules 1 to 5\label{tab:table1}}
\centering
\begin{tabular}{|c||c||c||c||c||c|}
\hline
  & Module 1 & Module 2 & Module 3 & Module 4 & Module 5\\
\hline
$k$ & 256 & 128 & 64 & 32 & 32\\
$m$ & 8 & 16 & 32 & 32 & 1\\
$z$ & 32 & 64 & 128 & 256 & 512\\
$e_1$ & 32 & 32 & 32 & 32 & 32\\
$e_2$ & 64 & 64 & 64 & 64 & 64\\
\hline
\end{tabular}
\end{table}

In Module 1, we set the number of downsampled LGAs to $k (k=256)$, and each LGA considers the topology information from $m (m=8)$ neighboring contour points. In this paper,  the number of channels for the two-layer multi-channel convolutions in the feature enhancement module is $e_1(e_1=32)$ and $e_2(e_2=64)$ respectively. Therefore, after feature alignment, the enhanced feature $X_1$ has a feature dimension of 64. Then, $X_1$ is compressed to $z(z=32)$ dimensions by the feature fusion module and sent to the next encoding module after max-pooling. In the subsequent modules, the contour shape is encoded through progressive down-sampling to achieve learning of more abstract geometric features. At the same time, during the encoding process, the feature output of each module is concatenated to the next moudule through feature fusion. The hyperparameter settings for each module are shown in Table I. It is worth noting that Encoding Module 5 is special. Inspired by RIConv++\cite{zhang2022riconv2}, we regard all $k$ sampled points as one LGA for encoding to capture the global features of the shape. In the feature fusion between different encoding modules, we ensure the consistency of feature dimensions through a filter, which is achieved by the reshape and repeat operations of tensors. Finally, the encoded features pass through three fully-connected layers and a softmax layer for classification.

\section{Experimental Results}

In this section, we design two tasks to validate the proposed encoding module. The first one is classification, which we test on two datasets: Kimia99\cite{kimia99} and Flavia\cite{flavia}. We compare the proposed module with recent studies on shape rotation-invariant recognition. We test the rotation-invariant recognition ability under arbitrary rotation angles and discuss its robustness under different numbers of augmented samples. The second is retrieval, where we added two new handcrafted datasets: Kidney and Liver. Their data are cross-sections from different locations of the liver and kidney, respectively. Organ cross-section retrieval is an important research direction in medical image registration\cite{regis2016,regis2019,regis2024}. In this task, we test the rotational invariance at arbitrary rotation angles and we set up an anti-noise test.

\subsection{DATASET}

All the datasets used in this paper are from public sources. Here, we provide detailed information about four datasets. The image-format data of all shapes is prepared by binarization, and the contour data is obtained from the automatic boundary extraction of the binarized images.In addition, the shape diagrams corresponding to the shape IDs and category IDs used in experiments are provided in the appendix.

\begin{figure}[h]
\centering
\includegraphics[width=8.8cm]{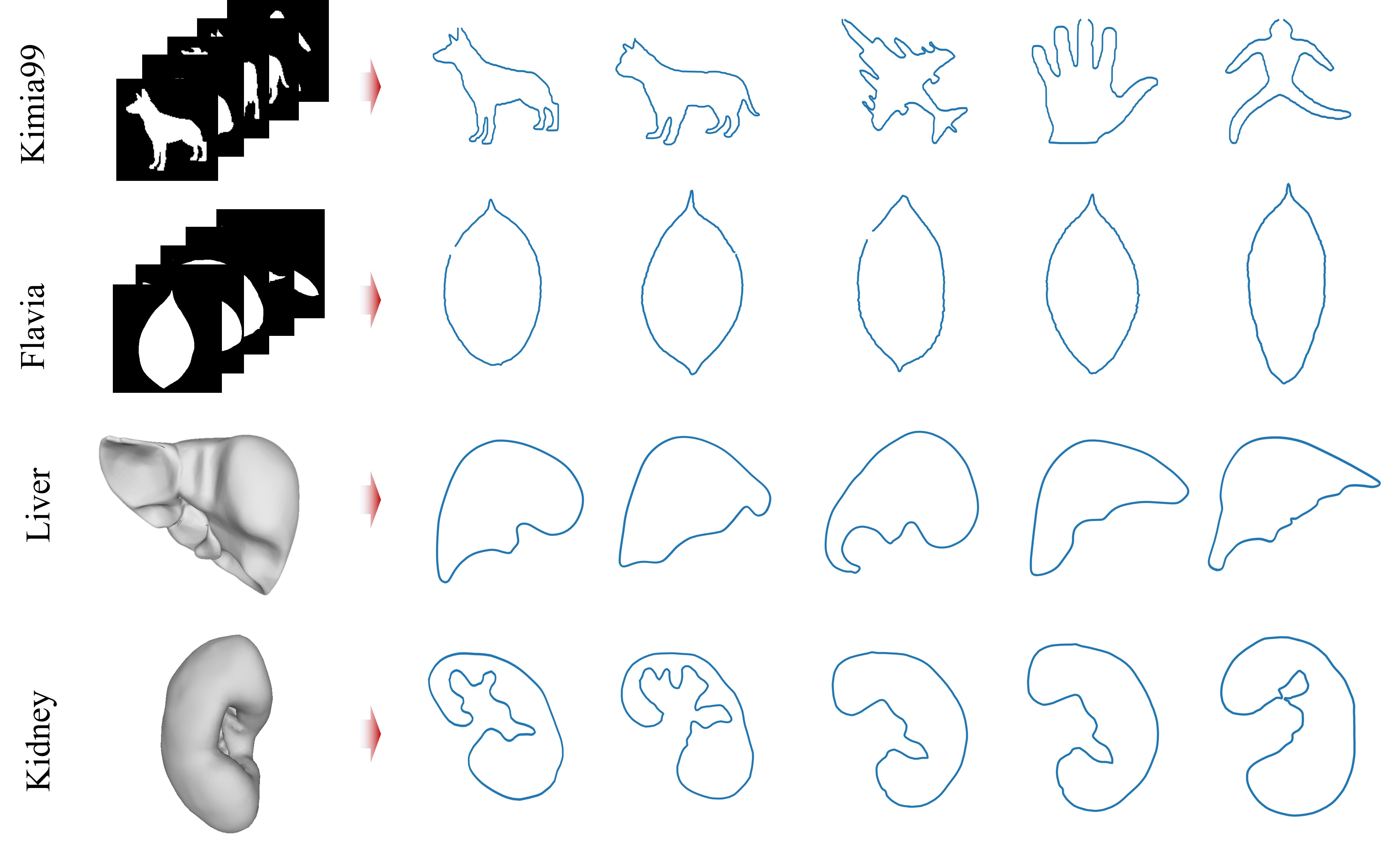}
\caption{Some samples from the datasets: Kimia99, Flavia, Liver and Kidney.}
\label{fig_7}
\end{figure}

\subsubsection{Kimia99}
Kimia99\cite{kimia99} consists of 9 shape categories with large inter-class differences. There are 11 shapes in each category, with variations in form, occlusion, articulation, missing parts, and so on. For the classification task, we select 80 percent of the data as the training sets and 20 percent as the testing sets. For the retrieval task, we randomly choose 19 shapes to form the dataset.

\subsubsection{Flavia}
Flavia\cite{flavia} involves the images of various plant leaves, which contains 1907 leaf images from 32 different plant species. As shown in Fig.\ref{fig_7}, the inter-class similarity of the Flavia dataset is quite high. Similarly, for the classification task, we select 80 percent of the data as the training set and 20 percent as the testing set. In the retrieval task, we randomly select 50 shapes as the dataset.

\subsubsection{Kidney}
The Kidney dataset consists of 146 cross-sectional contours from the 3D reconstructed kidney organs. The local structure of the contours is complex. Meanwhile, the cross-sectional differences at different positions are relatively small. It is a handcrafted dataset, while the kidney models are sourced from the open dataset AMOS\cite{amos2022}. This dataset is used for testing in the retrieval task. 

\subsubsection{Liver}
The Liver dataset contains 400 cross-sectional contours of liver organ models. The liver organ models are reconstructed from the public dataset BodyParts3D\cite{bodyparts3d2008}. The local geometric features of the cross-sections of this model are simple, and the similarity between different cross-sections is extremely high. This dataset is used for testing in the retrieval task.

\subsection{Implementation Details}

All comparisons and tests are conducted based on the network architecture illustrated in Fig.\ref{fig_6}. We implement our experiments in Python with the Pytorch on an NVIDIA GTX4090 GPU. In training stage, we use the Adam optimizer with a weight decay of 0.0001 and a learning rate of 0.001. We train the model for 300 epochs and use the results as the basis for testing. We compare our method with the latest research methods on rotation-invariant recognition(RIC-CNN\cite{riccnn2024},CRSA\cite{crsa2024},E2-CNN\cite{e2cnn2021},TIGCN\cite{tigcn2020},H-Net\cite{hnet2017},GCN\cite{gcn2018}). All comparative methods are executed under the provided source code.

\subsection{Shape Classification Tasks}

To verify the effectiveness of this module, we test the classification performance of the network framework on a coarse-grained dataset Kimia99 and a fine-grained dataset Flavia respectively. On the Kimia99 dataset, we randomly selected 3 shapes from each class for testing and used the remaining 9 shapes for training. Since there are 11 shape classes in total, the testing set contains 33 shapes and the training set contains 88 shapes. For the Flavia dataset, as the number of leaves in each class is inconsistent, we divide each shape category into a training set and a testing set at a ratio of 8:2. 

\subsubsection{Classification Performance Comparison}

\begin{figure}[h]
\centering
\includegraphics[width=8.8cm]{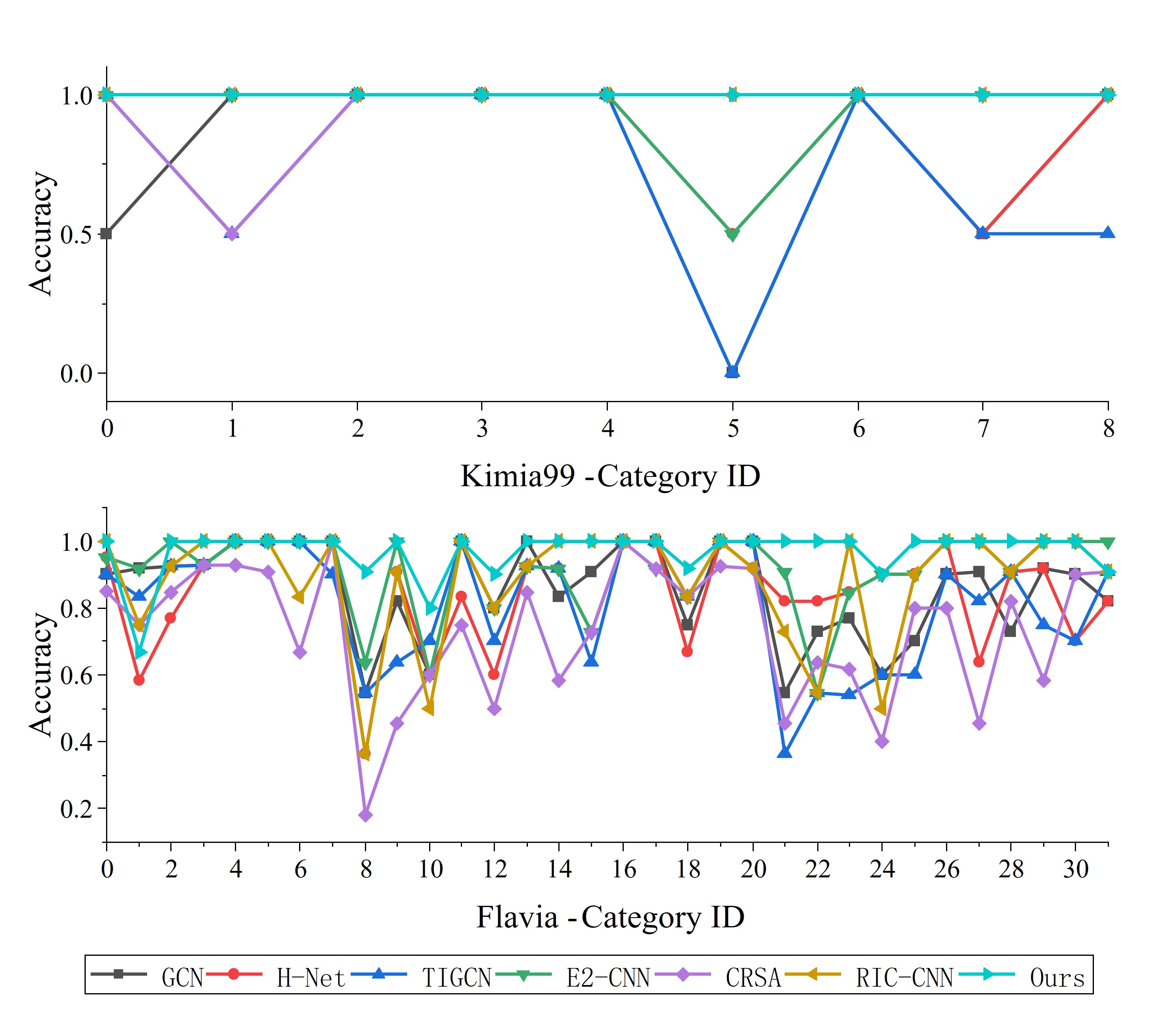}
\caption{The recognition accuracy of different shape categories in the shape classification task.}
\label{fig_8}
\end{figure}

We first conduct classification tests on two datasets representing coarse-grained and fine-grained tasks using various methods, without comparing the rotation invariance in this comparison. To compare the classification performance of different methods on shapes with varying geometric features across the two datasets, we illustrate the recognition accuracy of different categories in both datasets in Fig.\ref{fig_8}.

In the Kimia99 dataset, shapes in Category ID-7 exhibits rotation and partial occlusion defects, where HNet, GCN, and TIGCN fail to handle robustly. In addition to the aforementioned defects, E2-CNN also struggles with the large-scale missing regions and artifacts present in Category ID-5. According to the experimental results, our method, RIC-CNN and CRSA demonstrate higher robustness against these defects. In the Flavia dataset, the inter-class similarity is relatively low, making it difficult for pixel-patch-based deep learning methods to capture suitable clustering features. In contrast, our deep geometric encoding approach for shape contours is capable of learning more abstract classification features, enabling the differentiation of fine-grained shapes. Table 2 presents the average recognition accuracies of the listed methods on the two datasets. The results demonstrate that our method achieves an improvement of nearly $5\%$ in both fine-grained and coarse-grained shape classification tasks.

\begin{table}[h]
\caption{The average values of rotation-invariant recognition accuracy at any angle\label{tab:table2}}
\centering
\begin{tabular}{|c||c||c||c|}
\hline
Method & Data Size & Kimia99 & Flavia\\
\hline
GCN     & 32$\times$32   &77.78  &86.58 \\
H-Net   & 32$\times$32   &88.89  &85.00 \\
TIGCN   & 32$\times$32   &72.22  &82.37 \\
E2-CNN  & 32$\times$32   &94.44  &92.11 \\
CRSA    & 100$\times$100 &94.44  &75.79 \\
RIC-CNN & 32$\times$32   &100.00 &88.68 \\
Ours    & 320$\times$2   &100.00 &97.11 \\
\hline
\end{tabular}
\end{table}

\subsubsection{Rotational invariance in different number of sample enhancements}

\begin{figure*}[!t]
\centering
\includegraphics[width=17.2cm]{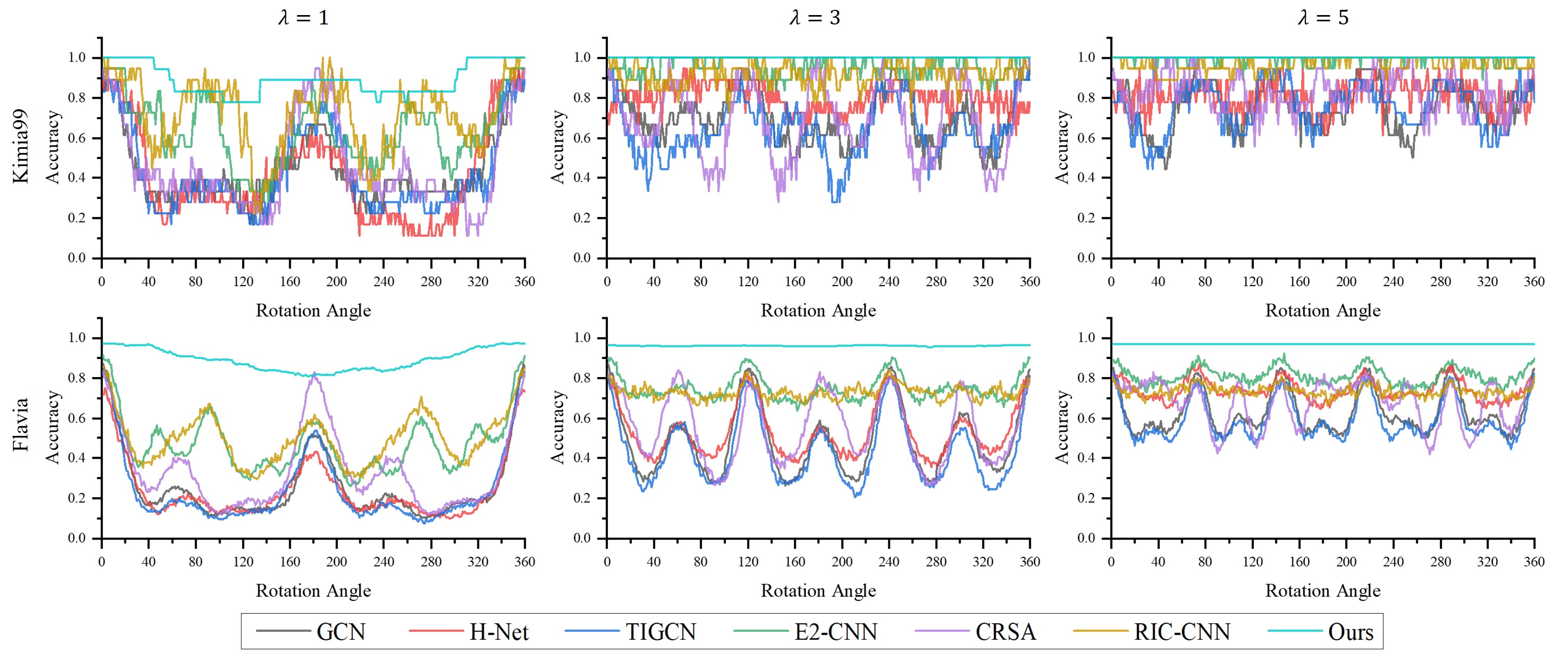}
\caption{Classification success rate at any rotation angle. We perform rotation augmentation on the training set at intervals of $360/\lambda$ within the range of $[0^\circ,360^\circ)$. Here, $\lambda=i$ represents the testing results using the training weights obtained after the corresponding rotation augmentation.}
\label{fig_9}
\end{figure*}

Furthermore, we test the rotation-invariant classification capability of the methods under arbitrary rotation angles. We introduce a parameter $\lambda$ to represent the degree of data augmentation applied to the training set. Specifically, $\lambda=i$ indicates that for each shape in the training set, we sample rotated versions at intervals of $360/\lambda^\circ$ as augmented training data. $\lambda=1$ denotes no data augmentation. For the test set, we sample rotated versions of each shape at $1^\circ$ intervals as additional test data. Based on the aforementioned training and test data, we compared the rotation-invariant recognition performance under varying degrees of data augmentation. Fig.\ref{fig_9} illustrates the recognition rates of the test set at different rotation angles.

\begin{table}[h]
\caption{The average rotation-invariant recognition accuracy with different numbers of data augmentations \label{tab:table3}}
\centering
\scalebox{0.9}{
\begin{tabular}{|c||c||c||c||c||c||c||c|}
\hline
\multirow{2}*{Method} & \multicolumn{3}{|c|}{Kimia99} & \multicolumn{3}{|c|}{Flavia}\\
\cline{2-7}
~ & $\lambda=1$ & $\lambda=3$ & $\lambda=5$& $\lambda=1$ & $\lambda=3$ & $\lambda=5$\\
\hline
GCN     & 43.75 & 69.72 & 75.00 & 25.50 & 48.75 & 62.79\\
H-Net   & 50.54 & 79.81 & 80.82 & 23.77 & 53.08 & 73.87 \\
TIGCN   & 42.81 & 63.86 & 75.60 & 23.88 & 43.03 & 58.97\\
E2-CNN  & 64.71 & 95.54 & 99.12 & 46.69 & 74.25 & 81.10 \\
CRSA    & 46.59 & 69.15 & 82.41 & 33.55 & 54.83 & 67.42\\
RIC-CNN & 71.59 & 91.48 & 96.64 & 49.63 & 72.84 & 73.42\\
Ours    & 89.15 & 100.00 & 100.00 & 89.14 & 95.96 & 96.58\\
\hline
\end{tabular}
}
\end{table}

In the absence of data augmentation($\lambda=1$), we can observe that E2-CNN, RIC-CNN, and our method achieve high rotation-invariant recognition at arbitrary angles. However, for the fine-grained shape dataset Flavia, methods such as CRIC-CNN and E2-GCN exhibit strong robustness only at specific angles($0^\circ, 90^\circ, 180^\circ, 270^\circ$), whereas our method maintains stable recognition across arbitrary angles. With the intervention  of data augmentation, when $\lambda=2$, E2-CNN and RIC-CNN have basically achieved rotation-invariant recognition at arbitrary angles, with their accuracy approaching the limits shown in Table II. When $\lambda=5$, H-Net also largely attains rotation-invariant recognition at arbitrary angles. In contrast, our method converges more quickly and stably to the upper limit of recognition accuracy shown in Table II. Table III presents the mean recognition accuracy across different angles under various $\lambda$ values. The results demonstrate that, without data augmentation($\lambda=1$), our method improves rotation-invariant classification accuracy by $17\%$ and $39\%$ for coarse-grained and fine-grained datasets, respectively. With the introduction of minimal data augmentation($\lambda=5$), the accuracy improvements are approximately $3\%$ and $15\%$, respectively.

\subsection{Shape Retrieval Tasks}

To further verify the recognition ability of this module, we conduct rotation-invariant retrieval tests on four datasets. Shape retrieval tasks are often used in applications such as face recognition, pose matching, and medical image registration. We make the following adjustments to the training data. In Kimia99, we use all 99 shapes as 99 categories for training. In Flavia, we select one shape from each leaf category, resulting in a total of 32 shape categories for training. Additionally, we introduce the Liver and Kidney datasets. In the retrieval task, we consider datasets Kimia99 and Flavia as coarse-grained datasets, while regarding Liver and Kidney as fine-grained datasets.

\subsubsection{Rotation-invariant retrieval at arbitrary angles}

\begin{figure*}[!h]
\centering
\includegraphics[width=17.2cm]{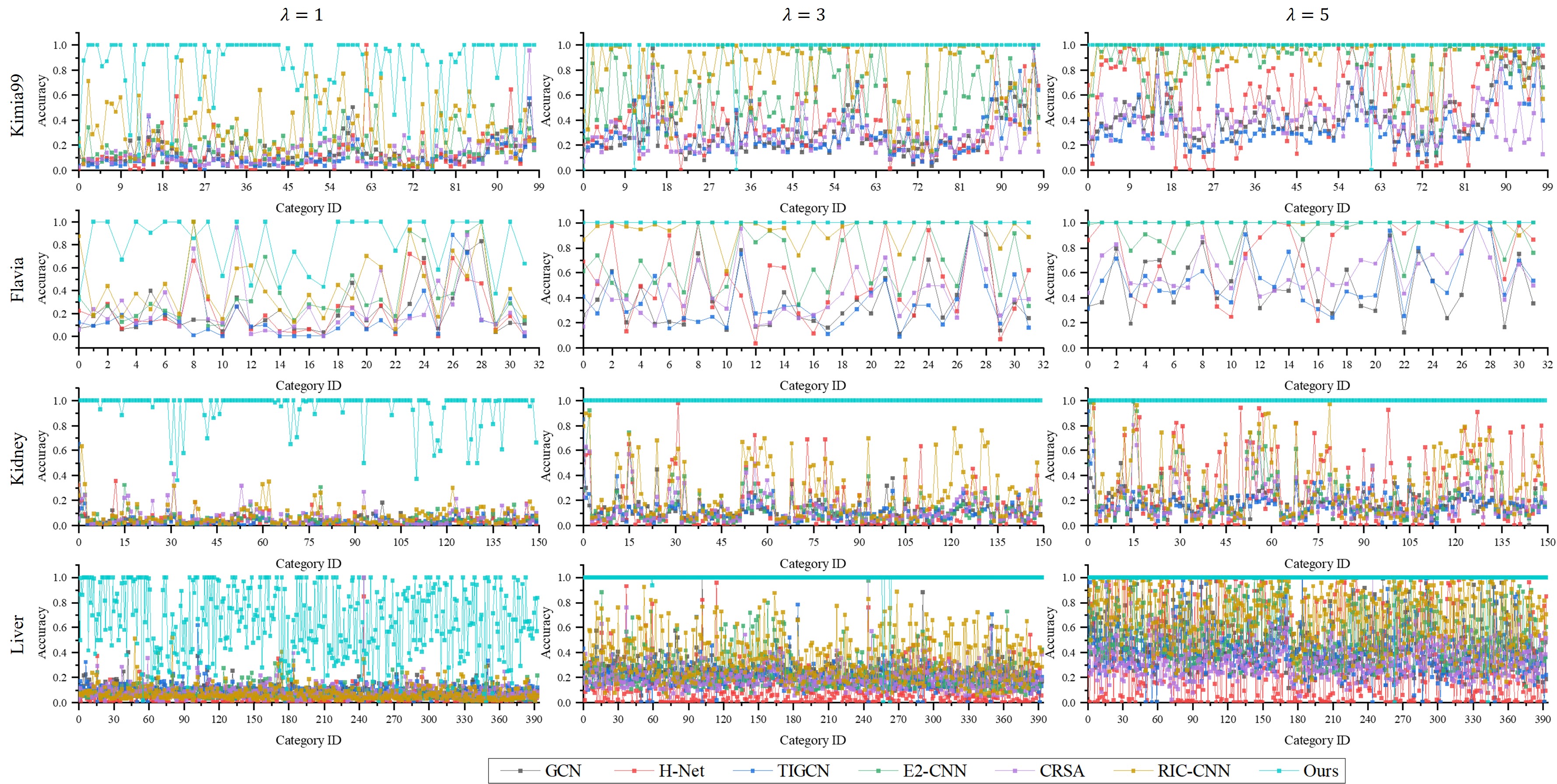}
\caption{The retrieval recognition accuracy of different shape categories.}
\label{fig_10}
\end{figure*}

First, we test the rotation-invariant retrieval performance at any angle. For the test data, we rotate all the shapes used for training once every 1 degree, a total of 360 times, and use these as the testsets to verify the rotation-invariant retrieval ability of the model. Fig.\ref{fig_10} shows the retrieval recognition accuracy of different methods at any angle under the four datasets. 

From the results of the retrieval task, our method achieves high success rates for rotation-invariant recognition at arbitrary angles on the four datasets, even without rotation data augmentation($\lambda=1$). The shapes with the lowest recognition accuracy in these datasets primarily belong to categories with extremely high inter-class similarity. However, for successfully recognized shapes, the rotation-invariant recognition accuracy at arbitrary angles is nearly $99\%$, a level that pixel-based methods struggle to achieve. After incorporating rotation-augmented datasets into the training process($\lambda=5$), our method attains a recognition accuracy of about $99\%$ for arbitrary-angle rotation invariance on the four datasets. Table \ref{tab:table4} presents the average retrieval accuracy of the tested methods across the four datasets under different data-augmented training sets. Compared to state-of-the-art pixel-based methods, our approach improves accuracy by $8\%$, $0.3\%$, $66\%$, and $30\%$ on the Kimia99, Flavia, Kidney, and Liver datasets, respectively.

\begin{table*}[h]
\caption{The average rotation-invariant retrieval accuracy with different numbers of data augmentations \label{tab:table4}}
\centering
\scalebox{1.0}{
\begin{tabular}{|c||c||c||c||c||c||c||c||c||c||c||c||c||c|}
\hline
\multirow{2}*{Method} & \multicolumn{3}{|c|}{Kimia99} & \multicolumn{3}{|c|}{Flavia} & \multicolumn{3}{|c|}{Kidney} & \multicolumn{3}{|c|}{Liver}\\
\cline{2-13}
~ & $\lambda=1$ & $\lambda=3$ & $\lambda=5$& $\lambda=1$ & $\lambda=3$ & $\lambda=5$& $\lambda=1$ & $\lambda=3$ & $\lambda=5$& $\lambda=1$ & $\lambda=3$ & $\lambda=5$\\
\hline
GCN& 12.45 & 28.01 & 43.60 & 21.40 & 37.02 & 53.39 & 3.83 & 11.28 & 17.24 & 9.14 & 23.49 & 43.50 \\ 
H-Net& 11.82 & 35.16 & 59.92 & 21.78 & 53.16 & 82.60 & 3.16 & 13.20 & 31.44 & 4.45 & 11.86 & 33.41 \\ 
TIGCN& 9.84 & 29.12 & 36.33 & 15.13 & 34.64 & 55.09 & 2.60 & 10.46 & 16.87 & 9.07 & 21.19 & 42.08 \\ 
E2-CNN& 18.78 & 58.35 & 86.62 & 36.68 & 67.34 & 93.16 & 4.39 & 12.33 & 23.81 & 7.29 & 23.36 & 53.20 \\ 
CRSA& 12.84 & 26.33 & 41.00 & 23.83 & 41.95 & 61.06 & 6.03 & 12.85 & 19.32 & 8.21 & 20.10 & 32.93 \\ 
RIC-CNN& 31.82 & 76.74 & 90.68 & 42.37 & 94.31 & 99.66 & 7.59 & 26.65 & 33.67 & 7.47 & 35.61 & 69.13 \\ 
Ours& 83.94 & 96.14 & 98.99 & 81.75 & 100.00 & 100.00 & 94.20 & 100.00 & 100.00 & 65.07 & 99.48 & 99.49 \\ 
\hline
\end{tabular}
}
\end{table*}

\begin{figure}[h]
\centering
\includegraphics[width=8.8cm]{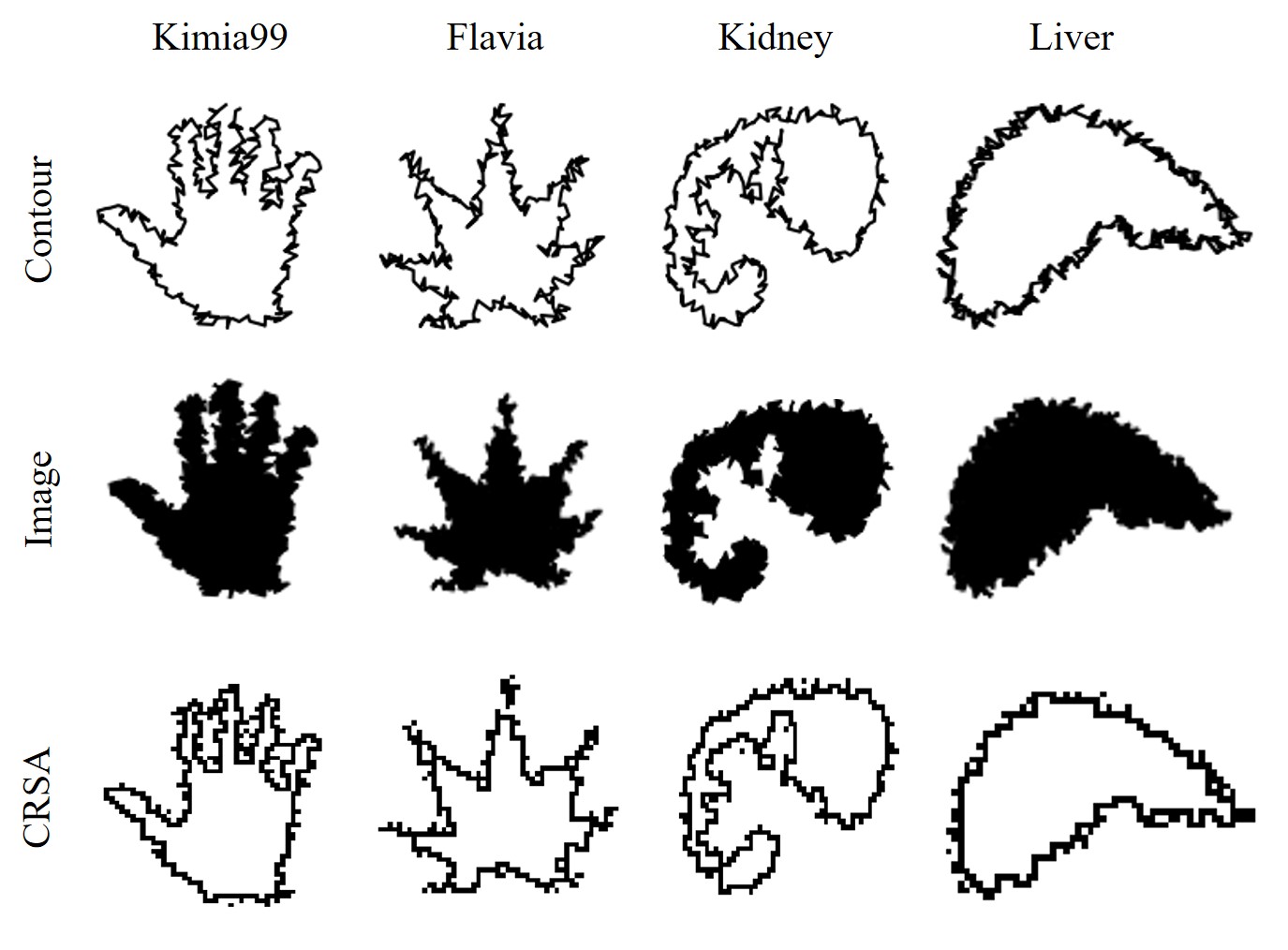}
\caption{The schematic representations of three data types for shapes with level-5 noise.}
\label{fig_11}
\end{figure}

\subsubsection{Robustness in different levels of contour noise}

\begin{figure}[h]
\centering
\includegraphics[width=8.8cm]{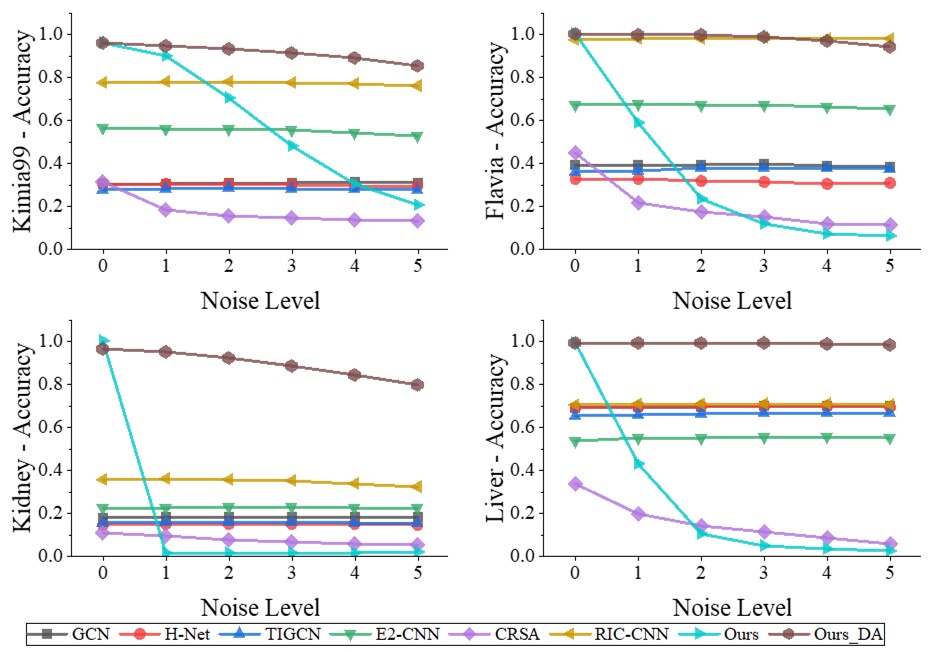}
\caption{Recognition accuracy for rotation at any angle under noise levels ranging from 0 to 5.}
\label{fig_12}
\end{figure}

In tasks such as face recognition or shape-matching, noise significantly impacts the recognition accuracy of neural networks. To examine the noise robustness of this module, based on the four datasets, we set six noise levels ranging from 0 to 5 according to their different scales. The level-5 noise of each shape is $1/24$ times the side length of its minimum bounding square. Fig.\ref{fig_11} illustrates the schematic representations of three data types for shapes with level-5 noise. In the noise test, for each shape on the six noise levels of the four datasets, we perform rotational sampling with a sampling interval of 5° to form the test set. We use the pre-trained weights of $\lambda=3$ obtained from the retrieval task for the test, and the results are shown in Fig.\ref{fig_12}. 

Compared with image-based methods, our method is less stable in noise-resistance robustness, especially in fine-grained shape datasets. This maybe because the encoding method based on point-topology information is highly sensitive to local geometric features, easily leading to over-fitting of shapes during the learning process of retrieval tasks. In contrast, image-based methods exhibit stronger robustness to boundary noise due to the leakage of geometric information at an early stage. The learning pattern sensitive to point-topology information is also reflected in the CRSA method, as the CRSA method is also established based on contour-topology information, yet it still does not go beyond the scope of image learning. This drawback is more pronounced in the Kidney dataset, which has the strongest inter-class similarity. Our solution is to add a small amount of noisy data augmentation during training to reduce the over-fitting phenomenon in retrieval learning. $Ours\_DA$ represents the test results obtained by adding three noisy shapes to each shape in the training dataset. During the testing process, we randomly generate boundary noise for each noise level and take the average of 10 groups. The results indicate that our method demonstrates good noise robustness after a small amount of data augmentation.

\begin{figure}[h]
\centering
\includegraphics[width=8.8cm]{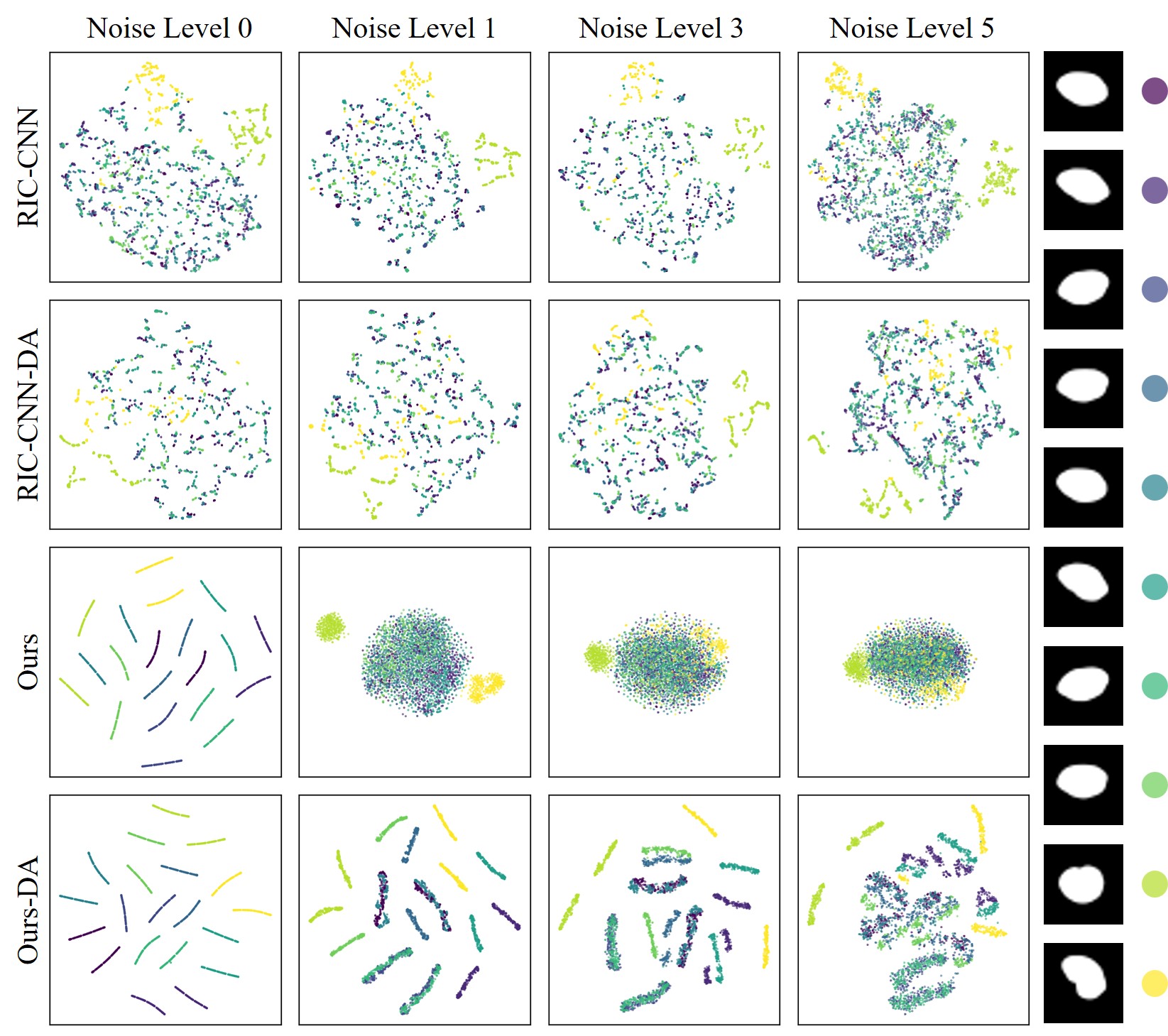}
\caption{Comparison of feature clustering capabilities for similar shapes. We compare the feature clustering capabilities of RIC-CNN and our method for similar shapes, and also compare the impacts of three levels of noise on them. }
\label{fig_13}
\end{figure}

In addition, to highlight the advantages of our method in rotation-invariant encoding of fine-grained shapes, we select the latest RIC-CNN method to compare with our method in terms of feature clustering ability. We select 10 fine-grained shapes from the Kidney dataset, which has the highest inter-class shape similarity. For the 10 shapes, we carry out training with $\lambda=3$ rotation augmentation. Subsequently, we conduct tests at noise levels of 0, 1, 3, and 5 with arbitrary angles. During the testing, we perform t-SNE feature clustering on the encoded features obtained from the layer preceding the network classification head. Fig.\ref{fig_13} shows the feature clustering of the rotation-encoded shapes within the range of $[0^\circ,360^\circ)$ for each shape under different noise levels. The results indicate that our method can achieve excellent identification of fine-grained shapes after introducing a small amount of noise augmentation.

\section{Discussion}

When storing an image with contour data, there are two types of rotation compared to the pixel images. The Type-I-Rotation is that the ordering of point sequence remains unchanged when the contour is rotated. The Type-II-Rotation is that the starting point of the point sequence changes, where this is manifested in Cartesian coordinates as a change in the phase of its coordinate arrangement. A common treatment for Type-II-Rotation is coordinate alignment\cite{Florian2021}. But because our method is the geometric feature encoding for closed curves, it is not sensitive to the coordinate changes caused by the two cases. Fig.\ref{fig_17} counts the recognition accuracies for the tests of Type-I, Type-II and Type-I+Type-II, where Type-I+Type-II refers to a random mixture of two rotation types. The results show that there is no significant difference in the recognition accuracy of the two cases for our model.

\begin{figure}[h]
\centering
\includegraphics[width=8.8cm]{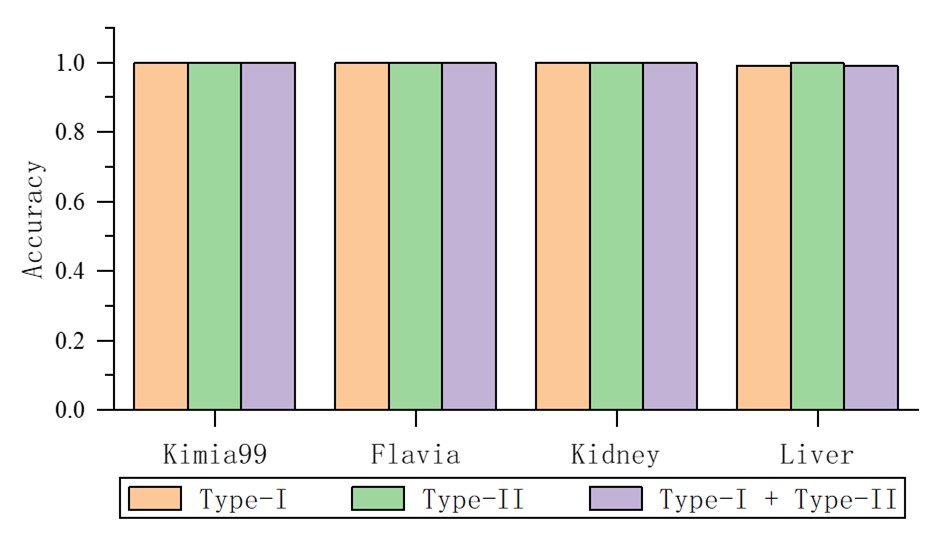}
\caption{Comparison of rotation-invariant recognition accuracy at any angle among the rotation types of Type-I, Type-II and Type-I + Type-II .}
\label{fig_14}
\end{figure}

\begin{figure*}[!t]
\centering
\includegraphics[width=17.2cm]{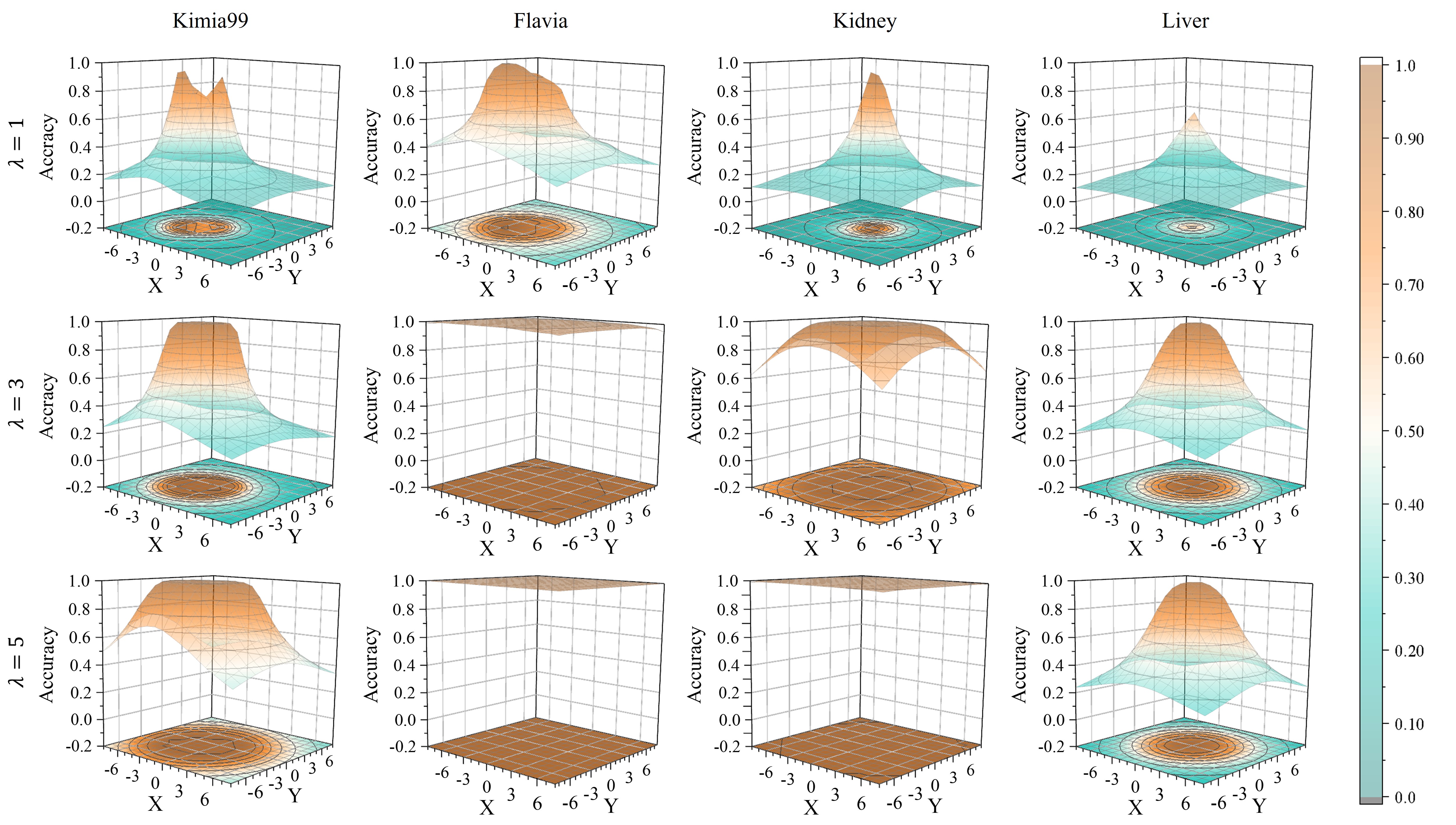}
\caption{Rotation-invariant recognition accuracy at any angle with different positions of the rotation center. The ranges of X and Y form a 17×17 mesh grid from $[-8,8]$. The mesh grid widths corresponding to the four datasets are $3\omega, 3\omega, \omega, \omega$ respectively, where $\omega$ is the maximum value from the side lengths of the minimum bounding squares.}
\label{fig_15}
\end{figure*}

In addition to rotation invariance, our method is also robust to the position of the rotation center. We set up a $17\times17$ mesh grid. We take the position of each grid point as the rotation center to calculate the rotation-invariant recognition accuracy at any angle. The results are shown in Fig.\ref{fig_15}, where the scales of X and Y represent the grid point numbers. The grid point $[0, 0]$ is the position of the shape centroid. In each dataset, we select the maximum value from the side lengths of the minimum bounding squares of all shapes and name it the parameter $\omega$. For the four datasets, we set the widths of mesh grid to $3\omega$, $3\omega$, $\omega$, $\omega$ respectively. Judging from the projection in Fig.\ref{fig_15}, our method can achieve relatively stable recognition within a circular area with the shape centroid as the origin. The radius of this circular area is roughly related to the granularity of the dataset. We can observe that for datasets with a high degree of granularity(Kimia99,Flavia), the radius is significantly larger than that of datasets with a low degree of granularity(Kidney,Liver). In addition, we found that rotation augmentation ($\lambda=3, \lambda=5$) of the training data can improve the robustness to the rotation position. However, this augmentation is limited, especially for datasets with a high degree of granularity.

In our approach, we proposed the LOA-based encoding, which is grounded in the contour direction. Within the framework of the proposed network, we compare the differences between the LRA and LOA. Specifically, we conduct comparative analyses on the Kidney and Kimia99 datasets. These two datasets are selected as representatives to explore the disparities in rotation-invariant recognition and noise resistance between the two methods. For rotation-invariant recognition, 

\begin{table*}[h]
\caption{The average retrieval success rates of LRA and LOA at any rotation angle with different numbers of data augmentation.\label{tab:table5}}
\centering
\scalebox{1.0}{
\begin{tabular}{|c||c||c||c||c||c||c||c||c||c||c||c|}
\hline
\multirow{2}*{Method} & \multicolumn{5}{|c|}{Kimia99} & \multicolumn{5}{|c|}{Kidney}\\
\cline{2-11}
~ & $\lambda=1$ & $\lambda=2$ & $\lambda=3$& $\lambda=4$ & $\lambda=5$ & $\lambda=1$& $\lambda=2$ & $\lambda=3$ & $\lambda=4$& $\lambda=5$\\
\hline
LRA& 18.39 & 25.17 & 91.57 & 96.06 & 98.29 & 6.65 & 10.35 & 30.02 & 26.90 & 80.09\\ 
LOA& 76.01 & 94.58 & 96.60 & 96.97 & 96.45 & 90.78 & 99.81 & 100.00 & 100.00 & 99.33\\ 
\hline
\end{tabular}
}
\end{table*}

\begin{figure}[!t]
\centering
\includegraphics[width=8.8cm]{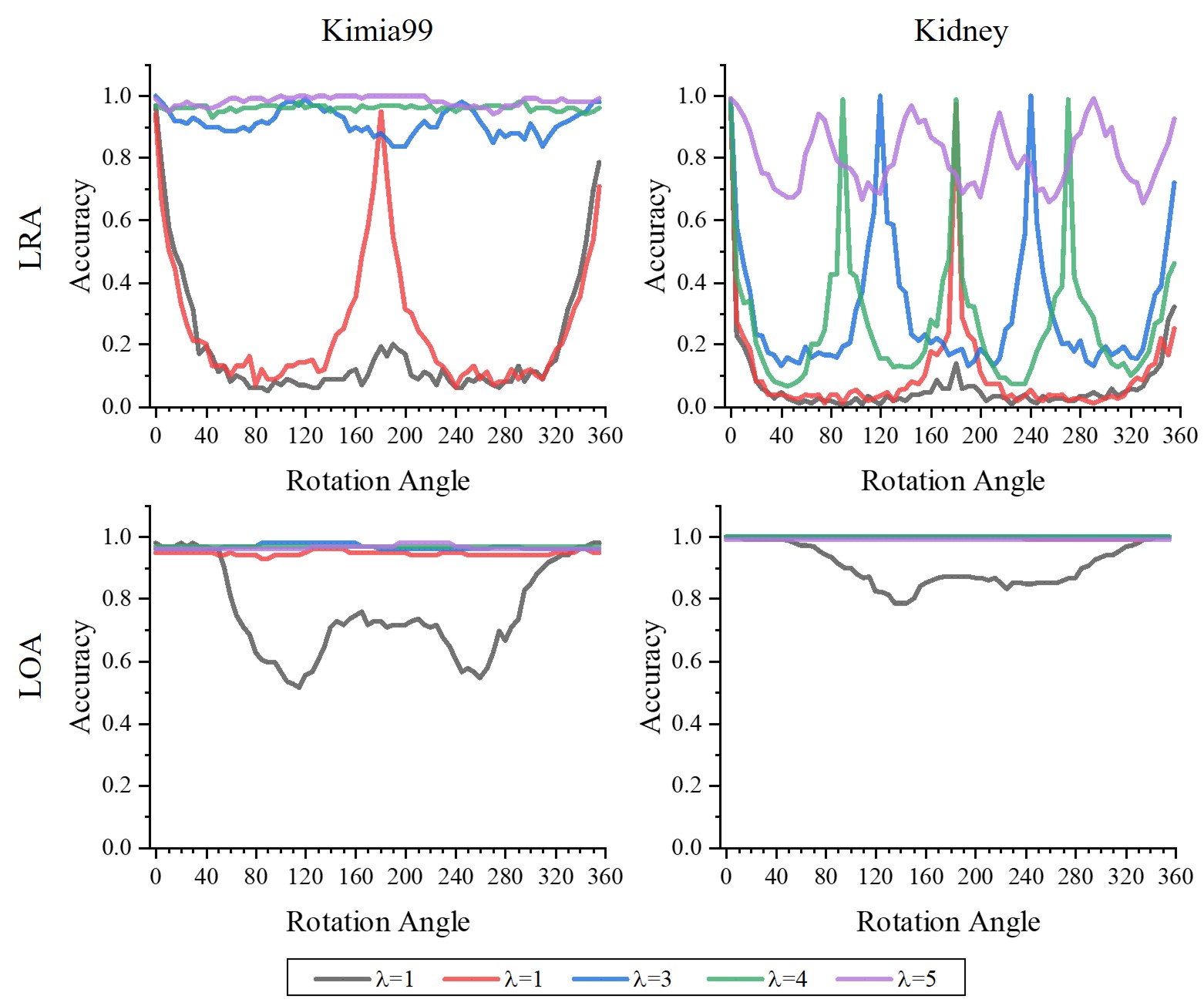}
\caption{Comparison of the rotation-invariant recognition accuracy at any angle between LOA and LRA under different numbers of augmented rotation samples.*$(\lambda=1,2,...,5)$}
\label{fig_16}
\end{figure}

\begin{figure}[!t]
\centering
\includegraphics[width=8.8cm]{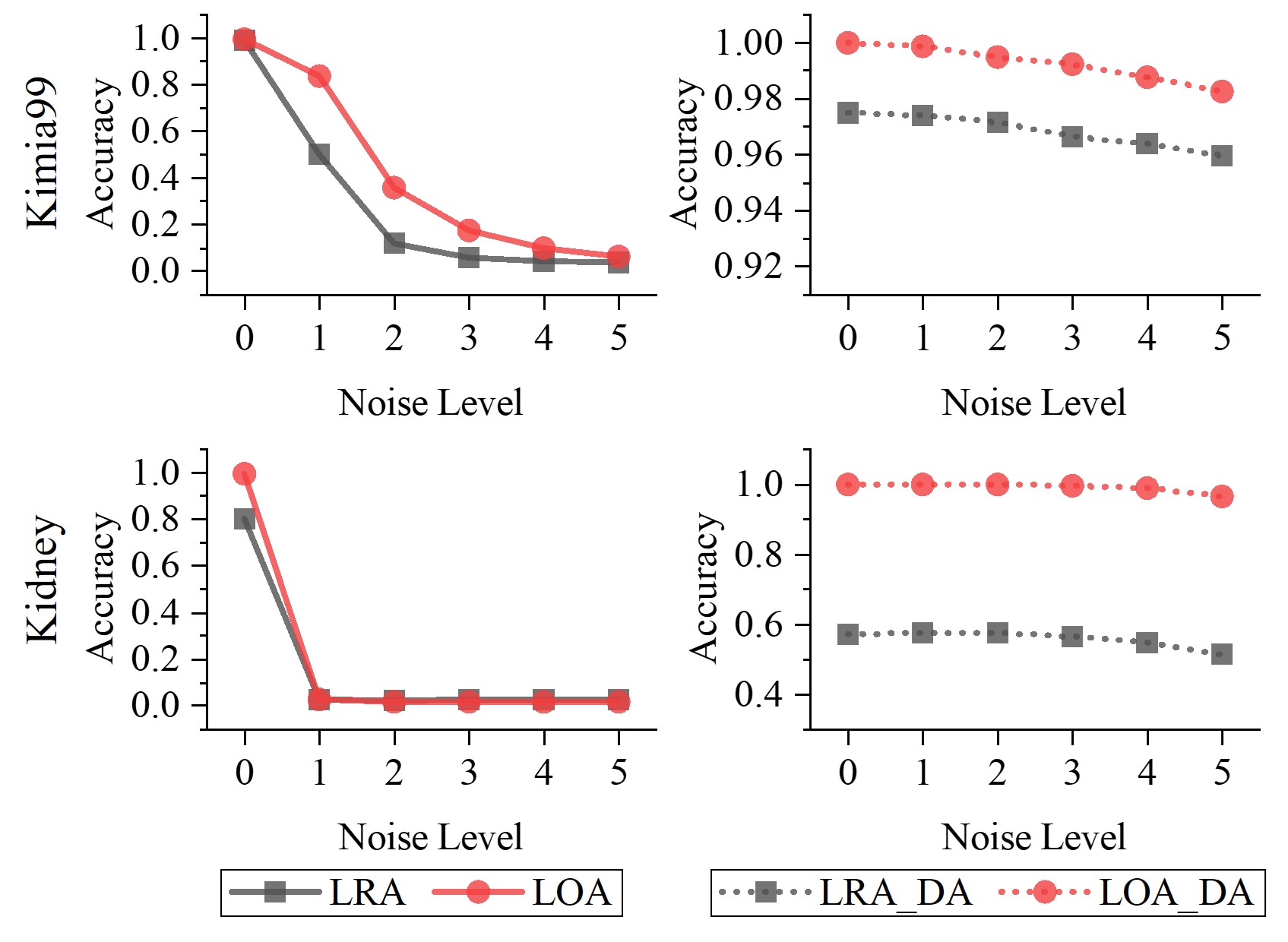}
\caption{Comparison of rotation-invariant recognition accuracy between LRA and LOA at different noise levels with and without noisy data augmentation.}
\label{fig_17}
\end{figure}

We carry out training processes under data augmentations with $\lambda=1,2,...,5$ respectively, and test the retrieval success rate of each shape at any rotation angle. The test results are illustrated in Fig.\ref{fig_16}. We can observe that, under no data augmentation ($\lambda=1$), the LRA method can only achieve a high accuracy near 0 degrees. Moreover, as the number of data augmentations increases, our method can converge more quickly. For the Kimia99 dataset, the LRA method tends to be stable when $\lambda=3$. While for the Kidney dataset, the LRA method still fails to achieve stable recognition when $\lambda=5$. However, LOA can achieve rapid convergence for both datasets after $\lambda=2$. Furthermore, we investigate the stability of LRA and LOA in terms of noise-resistance ability. For this purpose, we set up two groups of experiments, one with noise-free data augmentation and the other with noisy data augmentation. The results are shown in Fig.\ref{fig_17}. We can observe that under noise-free data augmentation, LOA is more robust to noise. Without considering the limit of recognition accuracy, LRA is more sensitive to noise than LOA, which is manifested in that the recognition accuracy of LRA drops more rapidly. After introducing noisy data augmentation, the degree of sensitivity is similar for both LRA and LOA.

\section{Conclusion}

Our aim is to rethink the rotation-invariant recognition of shapes from the perspective of contour points, with the expectation of enabling more accurate identification of fine-grained shapes. In this paper, we propose an anti-noise rotation-invariant convolution module based on contour geometric aware for fine-grained shape. The module has a stable ability to recognize rotated shapes, especially in the case of fine-grained shapes. We make inferences about shapes from the perspective of contour points rather than in the form of pixel image, which allows us to focus more on the geometric features of the shape. We propose a more robust local coding approach by attenuating the ambiguity of point-cloud normal vector. And the module can be easily embedded into neural networks. We design a large number of experiments for two tasks, shape classification and retrieval, to verify the advantages of the proposed method in achieving rotation-invariant recognition robustly. We compare six other recently proposed methods. The results show that our model has better performance in recognition and noise immunity for similar shape recognition. Finally, we provide a specific discussion of the performance and application of our model. We compare the recognition of contour data faced with two forms of rotation. Moreover, we verify that our method is also robust to the position of the rotation center, which is difficult to achieve even in the latest methods. And, we discuss the differences in model training caused by computing features via LOA and LRA. In the future, we hope that our method can have an impact in broader fields. Besides encoding the contour information of images, this module can be applied to achieve rotation-invariant recognition and analysis of any 2D signals and other data. Meanwhile, we also hope that this work can be applied to more specific tasks such as shape matching, face recognition, or position estimation. Currently the model needs to ensure data ordering of the input contour points, and we will break this limitation in the future.

\section*{Acknowledgments}
The authors  appreciate the financial support provided by the National Key Research and Development Program of China under Grant 2022YFB4703000 and are thankful for the reviews’ constructive comments.

\section{Appendix}
The contents in Appendix are provided as Supplemental Materials documents.



\bibliographystyle{IEEEtran}
\bibliography{IEEEabrv,Ref}

\begin{thebibliography}{10}
\providecommand{\url}[1]{#1}
\csname url@samestyle\endcsname
\providecommand{\newblock}{\relax}
\providecommand{\bibinfo}[2]{#2}
\providecommand{\BIBentrySTDinterwordspacing}{\spaceskip=0pt\relax}
\providecommand{\BIBentryALTinterwordstretchfactor}{4}
\providecommand{\BIBentryALTinterwordspacing}{\spaceskip=\fontdimen2\font plus
\BIBentryALTinterwordstretchfactor\fontdimen3\font minus \fontdimen4\font\relax}
\providecommand{\BIBforeignlanguage}[2]{{%
\expandafter\ifx\csname l@#1\endcsname\relax
\typeout{** WARNING: IEEEtran.bst: No hyphenation pattern has been}%
\typeout{** loaded for the language `#1'. Using the pattern for}%
\typeout{** the default language instead.}%
\else
\language=\csname l@#1\endcsname
\fi
#2}}
\providecommand{\BIBdecl}{\relax}
\BIBdecl

\bibitem{Attneave1954}
F.~Attneave, ``Some informational aspects of visual perception.'' \emph{Psychological review}, vol. 61 3, pp. 183--93, 1954.

\bibitem{Wei2023}
X.~Wei, R.~Yu, and J.~Sun, ``Learning view-based graph convolutional network for multi-view 3d shape analysis,'' \emph{IEEE Transactions on Pattern Analysis and Machine Intelligence}, vol.~45, no.~6, pp. 7525--7541, 2023.

\bibitem{Taeyeop2021}
T.~Lee, B.-U. Lee, M.~Kim, and I.~S. Kweon, ``Category-level metric scale object shape and pose estimation,'' \emph{IEEE Robotics and Automation Letters}, vol.~6, no.~4, pp. 8575--8582, 2021.

\bibitem{Li2021}
Z.~Li, A.~Heyden, and M.~Oskarsson, ``A novel joint points and silhouette-based method to estimate 3d human pose and shape,'' in \emph{Pattern Recognition. ICPR International Workshops and Challenges}, 2021.

\bibitem{Florian2021}
F.~Lardeux, S.~Marchand, and P.~Gomez-Krämer, ``Low-complexity arrays of contour signatures for exact shape retrieval,'' \emph{Pattern Recognition}, vol. 118, p. 108000, 2021.

\bibitem{Bansal2020}
M.~Bansal, M.~Kumar, and M.~Kumar, ``2d object recognition techniques: State-of-the-art work,'' \emph{Archives of Computational Methods in Engineering}, vol.~28, pp. 1147 -- 1161, 2020.

\bibitem{Mei2023}
S.~Mei, R.~Jiang, M.~Ma, and C.~Song, ``Rotation-invariant feature learning via convolutional neural network with cyclic polar coordinates convolutional layer,'' \emph{IEEE Transactions on Geoscience and Remote Sensing}, vol.~61, pp. 1--13, 2023.

\bibitem{Zeiler2014}
M.~D. Zeiler and R.~Fergus, ``Visualizing and understanding convolutional networks,'' in \emph{Computer Vision -- ECCV 2014}.\hskip 1em plus 0.5em minus 0.4em\relax Springer International Publishing, 2014, pp. 818--833.

\bibitem{Bai2020}
X.~Bai, Z.~Luo, L.~Zhou, H.~Fu, L.~Quan, and C.-L. Tai, ``D3feat: Joint learning of dense detection and description of 3d local features,'' in \emph{2020 IEEE/CVF Conference on Computer Vision and Pattern Recognition (CVPR)}, 2020, pp. 6358--6366.

\bibitem{Bergmann2023}
P.~Bergmann and D.~Sattlegger, ``Anomaly detection in 3d point clouds using deep geometric descriptors,'' in \emph{2023 IEEE/CVF Winter Conference on Applications of Computer Vision (WACV)}, 2023, pp. 2612--2622.

\bibitem{jaderberg2015}
M.~Jaderberg, K.~Simonyan, A.~Zisserman \emph{et~al.}, ``Spatial transformer networks,'' \emph{Advances in neural information processing systems}, vol.~28, 2015.

\bibitem{Cohen2016}
T.~S. Cohen and M.~Welling, ``Group equivariant convolutional networks,'' in \emph{Proceedings of the 33rd International Conference on International Conference on Machine Learning}, vol.~48, 2016, p. 2990–2999.

\bibitem{Worrall2017}
D.~E. Worrall, S.~J. Garbin, D.~Turmukhambetov, and G.~J. Brostow, ``Harmonic networks: Deep translation and rotation equivariance,'' in \emph{2017 IEEE Conference on Computer Vision and Pattern Recognition (CVPR)}, 2017, pp. 7168--7177.

\bibitem{Lucas2024}
L.~C. Ribas and O.~M. Bruno, ``Learning a complex network representation for shape classification,'' \emph{Pattern Recognition}, vol. 154, p. 110566, 2024.

\bibitem{riccnn2024}
H.~Mo and G.~Zhao, ``Ric-cnn: Rotation-invariant coordinate convolutional neural network,'' \emph{Pattern Recognition}, vol. 146, p. 109994, 2024.

\bibitem{Mohseni2019}
S.~S. Mohseni~Salehi, S.~Khan, D.~Erdogmus, and A.~Gholipour, ``Real-time deep pose estimation with geodesic loss for image-to-template rigid registration,'' \emph{IEEE Transactions on Medical Imaging}, vol.~38, no.~2, pp. 470--481, 2019.

\bibitem{Hori2022}
R.~Hori, R.~Hachiuma, M.~Isogawa, D.~Mikami, and H.~Saito, ``Silhouette-based 3d human pose estimation using a single wrist-mounted 360° camera,'' \emph{IEEE Access}, vol.~10, pp. 54\,957--54\,968, 2022.

\bibitem{Bosma2023}
L.~S. Bosma, M.~Ries, B.~{Denis de Senneville}, B.~W. Raaymakers, and C.~Zachiu, ``Integration of operator-validated contours in deformable image registration for dose accumulation in radiotherapy,'' \emph{Physics and Imaging in Radiation Oncology}, vol.~27, p. 100483, 2023.

\bibitem{Chang2024}
L.-J. Chang, Y.-C. Liao, C.-H. Lin, S.-F. Yang-Mao, and H.-T. Chen, ``Estimating 3d hand poses and shapes from silhouettes,'' \emph{APSIPA Transactions on Signal and Information Processing}, vol.~13, no.~5, 2024.

\bibitem{zheng2024}
S.~Zheng, Y.~Wang, X.~Yang, X.~Deng, M.~Ding, and W.~Hou, ``Deep rigid registration for slice-to-volume in real time,'' \emph{Expert Systems with Applications}, vol. 235, p. 121132, 2024.

\bibitem{pointnet2017}
R.~Q. Charles, H.~Su, M.~Kaichun, and L.~J. Guibas, ``Pointnet: Deep learning on point sets for 3d classification and segmentation,'' in \emph{2017 IEEE Conference on Computer Vision and Pattern Recognition (CVPR)}, 2017, pp. 77--85.

\bibitem{Chen2022}
R.~Chen and Y.~Cong, ``The devil is in the pose: Ambiguity-free 3d rotation-invariant learning via pose-aware convolution,'' in \emph{2022 IEEE/CVF Conference on Computer Vision and Pattern Recognition (CVPR)}, 2022, pp. 7462--7471.

\bibitem{zhang2022riconv2}
Z.~Zhang, B.-S. Hua, and S.-K. Yeung, ``Riconv++: Effective rotation invariant convolutions for 3d point clouds deep learning,'' \emph{International Journal of Computer Vision}, vol.~1, pp. 1--16, 2022.

\bibitem{Ojala2002}
T.~Ojala, M.~Pietikainen, and T.~Maenpaa, ``Multiresolution gray-scale and rotation invariant texture classification with local binary patterns,'' \emph{IEEE Transactions on Pattern Analysis and Machine Intelligence}, vol.~24, no.~7, pp. 971--987, 2002.

\bibitem{Lazebnik2005}
S.~Lazebnik, C.~Schmid, and J.~Ponce, ``A sparse texture representation using local affine regions,'' \emph{IEEE Transactions on Pattern Analysis and Machine Intelligence}, vol.~27, no.~8, pp. 1265--1278, 2005.

\bibitem{Herbert2008}
H.~Bay, A.~Ess, T.~Tuytelaars, and L.~{Van Gool}, ``Speeded-up robust features (surf),'' \emph{Computer Vision and Image Understanding}, vol. 110, no.~3, pp. 346--359, 2008.

\bibitem{rift2004}
\BIBentryALTinterwordspacing
D.~G. Lowe, ``Distinctive image features from scale-invariant keypoints,'' vol.~60, no.~2, pp. 91--110. [Online]. Available: \url{https://doi.org/10.1023/B:VISI.0000029664.99615.94}
\BIBentrySTDinterwordspacing

\bibitem{Fan2012}
B.~Fan, F.~Wu, and Z.~Hu, ``Rotationally invariant descriptors using intensity order pooling,'' \emph{IEEE Transactions on Pattern Analysis and Machine Intelligence}, vol.~34, no.~10, pp. 2031--2045, 2012.

\bibitem{Finnveden2020}
L.~Finnveden, Y.~Jansson, and T.~Lindeberg, ``Understanding when spatial transformer networks do not support invariance, and what to do about it,'' in \emph{2020 25th International Conference on Pattern Recognition (ICPR)}, 2021, pp. 3427--3434.

\bibitem{Duan2017}
Y.~Duan, J.~Lu, J.~Feng, and J.~Zhou, ``Learning rotation-invariant local binary descriptor,'' \emph{IEEE Transactions on Image Processing}, vol.~26, no.~8, pp. 3636--3651, 2017.

\bibitem{Miao2021}
Y.~Miao, Z.~Lin, X.~Ma, G.~Ding, and J.~Han, ``Learning transformation-invariant local descriptors with low-coupling binary codes,'' \emph{IEEE Transactions on Image Processing}, vol.~30, pp. 7554--7566, 2021.

\bibitem{Hao2024}
\BIBentryALTinterwordspacing
H.~Wu, L.~Fang, Q.~Yu, and C.~Yang, ``Composite descriptor based on contour and appearance for plant species identification,'' \emph{Engineering Applications of Artificial Intelligence}, vol. 133, p. 108291, 2024. [Online]. Available: \url{https://www.sciencedirect.com/science/article/pii/S0952197624004494}
\BIBentrySTDinterwordspacing

\bibitem{Maurya2024}
\BIBentryALTinterwordspacing
A.~Maurya and D.~Singh, ``Rotation, scaling, and translation invariant an optimized and effective robust watermarking scheme,'' vol.~83, no.~7, pp. 20\,033--20\,053. [Online]. Available: \url{https://doi.org/10.1007/s11042-023-16321-w}
\BIBentrySTDinterwordspacing

\bibitem{Kaur2024}
\BIBentryALTinterwordspacing
M.~Kaur and S.~Vijay, ``Deep learning with invariant feature based species classification in underwater environments,'' vol.~83, no.~7, pp. 19\,587--19\,608. [Online]. Available: \url{https://doi.org/10.1007/s11042-023-15896-8}
\BIBentrySTDinterwordspacing

\bibitem{hnet2017}
D.~E. Worrall, S.~J. Garbin, D.~Turmukhambetov, and G.~J. Brostow, ``Harmonic networks: Deep translation and rotation equivariance,'' in \emph{2017 IEEE Conference on Computer Vision and Pattern Recognition (CVPR)}, 2017, pp. 7168--7177.

\bibitem{gcn2018}
S.~Luan, C.~Chen, B.~Zhang, J.~Han, and J.~Liu, ``Gabor convolutional networks,'' \emph{IEEE Transactions on Image Processing}, vol.~27, no.~9, pp. 4357--4366, 2018.

\bibitem{tigcn2020}
\BIBentryALTinterwordspacing
L.~Zhuang, F.~Da, S.~Gai, and M.~Li, ``Transformation-invariant gabor convolutional networks,'' vol.~14, no.~7, pp. 1413--1420. [Online]. Available: \url{https://doi.org/10.1007/s11760-020-01684-6}
\BIBentrySTDinterwordspacing

\bibitem{e2cnn2021}
\BIBentryALTinterwordspacing
M.~Weiler and G.~Cesa, ``General $e(2)$-equivariant steerable cnns,'' 2021. [Online]. Available: \url{https://arxiv.org/abs/1911.08251}
\BIBentrySTDinterwordspacing

\bibitem{Mo2024}
H.~Mo and G.~Zhao, ``Sorting convolution operation for achieving rotational invariance,'' \emph{IEEE Signal Processing Letters}, vol.~31, pp. 1199--1203, 2024.

\bibitem{Li2024}
H.~Li, R.~Pan, G.~Liu, M.~Dang, Q.~Xu, X.~Wang, and B.~Wan, ``Tir-net: Task integration based on rotated convolution kernel for oriented object detection in aerial images,'' \emph{IEEE Transactions on Geoscience and Remote Sensing}, vol.~62, pp. 1--13, 2024.

\bibitem{crsa2024}
L.~C. Ribas and O.~M. Bruno, ``Learning a complex network representation for shape classification,'' \emph{Pattern Recognition}, vol. 154, p. 110566, 2024.

\bibitem{Jia2024}
\BIBentryALTinterwordspacing
Q.~Jia, X.~Chen, Y.~Wang, X.~Fan, H.~Ling, and L.~J. Latecki, ``A rotation robust shape transformer for cartoon character recognition,'' vol.~40, no.~8, pp. 5575--5588. [Online]. Available: \url{https://doi.org/10.1007/s00371-023-03123-2}
\BIBentrySTDinterwordspacing

\bibitem{Kadam2022}
P.~Kadam, M.~Zhang, S.~Liu, and C.~C.~J. Kuo, ``R-pointhop: A green, accurate, and unsupervised point cloud registration method,'' \emph{IEEE Transactions on Image Processing}, vol.~31, pp. 2710--2725, 2022.

\bibitem{Rautek2024}
P.~Rautek, X.~Zhang, B.~Woschizka, T.~Theußl, and M.~Hadwiger, ``Vortex lens: Interactive vortex core line extraction using observed line integral convolution,'' \emph{IEEE Transactions on Visualization and Computer Graphics}, vol.~30, no.~1, pp. 55--65, 2024.

\bibitem{Fei2024}
J.~Fei and Z.~Deng, ``Rotation invariance and equivariance in 3d deep learning: a survey,'' \emph{Artif. Intell. Rev.}, vol.~57, p. 168, 2024.

\bibitem{fpfh2009rusu}
R.~B. Rusu, N.~Blodow, and M.~Beetz, ``Fast point feature histograms (fpfh) for 3d registration,'' in \emph{2009 IEEE International Conference on Robotics and Automation}, 2009, pp. 3212--3217.

\bibitem{shot2017samuele}
S.~Salti, F.~Tombari, and L.~{Di Stefano}, ``Shot: Unique signatures of histograms for surface and texture description,'' \emph{Computer Vision and Image Understanding}, vol. 125, pp. 251--264, 2014.

\bibitem{RoPS2013Guo}
Y.~Guo, F.~Sohel, M.~Bennamoun, M.~Lu, and J.~Wan, ``Rotational projection statistics for 3d local surface description and object recognition,'' vol. 105, no.~1, pp. 63--86.

\bibitem{kimia99}
T.~Sebastian, P.~Klein, and B.~Kimia, ``Recognition of shapes by editing their shock graphs,'' \emph{IEEE Transactions on Pattern Analysis and Machine Intelligence}, vol.~26, no.~5, pp. 550--571, 2004.

\bibitem{flavia}
S.~G. Wu, F.~S. Bao, E.~Y. Xu, Y.-X. Wang, Y.-F. Chang, and Q.-L. Xiang, ``A leaf recognition algorithm for plant classification using probabilistic neural network,'' in \emph{2007 IEEE International Symposium on Signal Processing and Information Technology}, 2007, pp. 11--16.

\bibitem{regis2016}
S.~Miao, Z.~J. Wang, and R.~Liao, ``A cnn regression approach for real-time 2d/3d registration,'' \emph{IEEE Transactions on Medical Imaging}, vol.~35, no.~5, pp. 1352--1363, 2016.

\bibitem{regis2019}
S.~S. Mohseni~Salehi, S.~Khan, D.~Erdogmus, and A.~Gholipour, ``Real-time deep pose estimation with geodesic loss for image-to-template rigid registration,'' \emph{IEEE Transactions on Medical Imaging}, vol.~38, no.~2, pp. 470--481, 2019.

\bibitem{regis2024}
F.~Liebmann, M.~{von Atzigen}, D.~Stütz, J.~Wolf, L.~Zingg, D.~Suter, N.~A. Cavalcanti, L.~Leoty, H.~Esfandiari, J.~G. Snedeker, M.~R. Oswald, M.~Pollefeys, M.~Farshad, and P.~Fürnstahl, ``Automatic registration with continuous pose updates for marker-less surgical navigation in spine surgery,'' \emph{Medical Image Analysis}, vol.~91, p. 103027, 2024.

\bibitem{amos2022}
Y.~Ji, H.~Bai, C.~Ge, J.~Yang, Y.~Zhu, R.~Zhang, Z.~Li, L.~Zhang, W.~Ma, X.~Wan, and P.~Luo, ``Amos: a large-scale abdominal multi-organ benchmark for versatile medical image segmentation,'' in \emph{Proceedings of the 36th International Conference on Neural Information Processing Systems}, ser. NIPS '22, 2024.

\bibitem{bodyparts3d2008}
N.~Mitsuhashi, K.~Fujieda, T.~Tamura, S.~Kawamoto, T.~Takagi, and K.~Okubo, ``Bodyparts3d: 3d structure database for anatomical concepts,'' \emph{Nucleic Acids Research}, vol.~37, no.~1, pp. D782--D785, 10 2008.

\end{thebibliography}




\vfill

\end{document}